\documentclass[10pt,twocolumn,letterpaper]{article}

\usepackage[pagenumbers]{iccv} 
\usepackage{marvosym}

\usepackage{graphicx}%
\usepackage{multirow}%
\usepackage{amsmath,amssymb,amsfonts}%
\usepackage{amsthm}%
\usepackage{mathrsfs}%
\usepackage[title]{appendix}%
\usepackage{xcolor}%
\usepackage{textcomp}%
\usepackage{manyfoot}%
\usepackage{booktabs}%
\usepackage{algorithm}%
\usepackage{algorithmicx}%
\usepackage{algpseudocode}%
\usepackage{listings}%
\usepackage{enumitem}
\usepackage{paralist}
\usepackage{capt-of}
\theoremstyle{thmstyleone}%
\definecolor{steponecircle}{RGB}{0,163,242}
\definecolor{steptwocircle}{RGB}{188,42,234}
\definecolor{stepthreecircle}{RGB}{233,113,50}
\definecolor{stepfourcircle}{RGB}{18,182,92}

\definecolor{customcolor}{HTML}{5D6EBA}

\usepackage{paralist}
\newcommand{\paratitle}[1]{\vspace{1.2ex}\noindent \textbf{#1}}

\usepackage{xspace}
\usepackage{pifont}

\newcommand{\DrVLogo}{\raisebox{-6pt}{\includegraphics[width=2.2em]{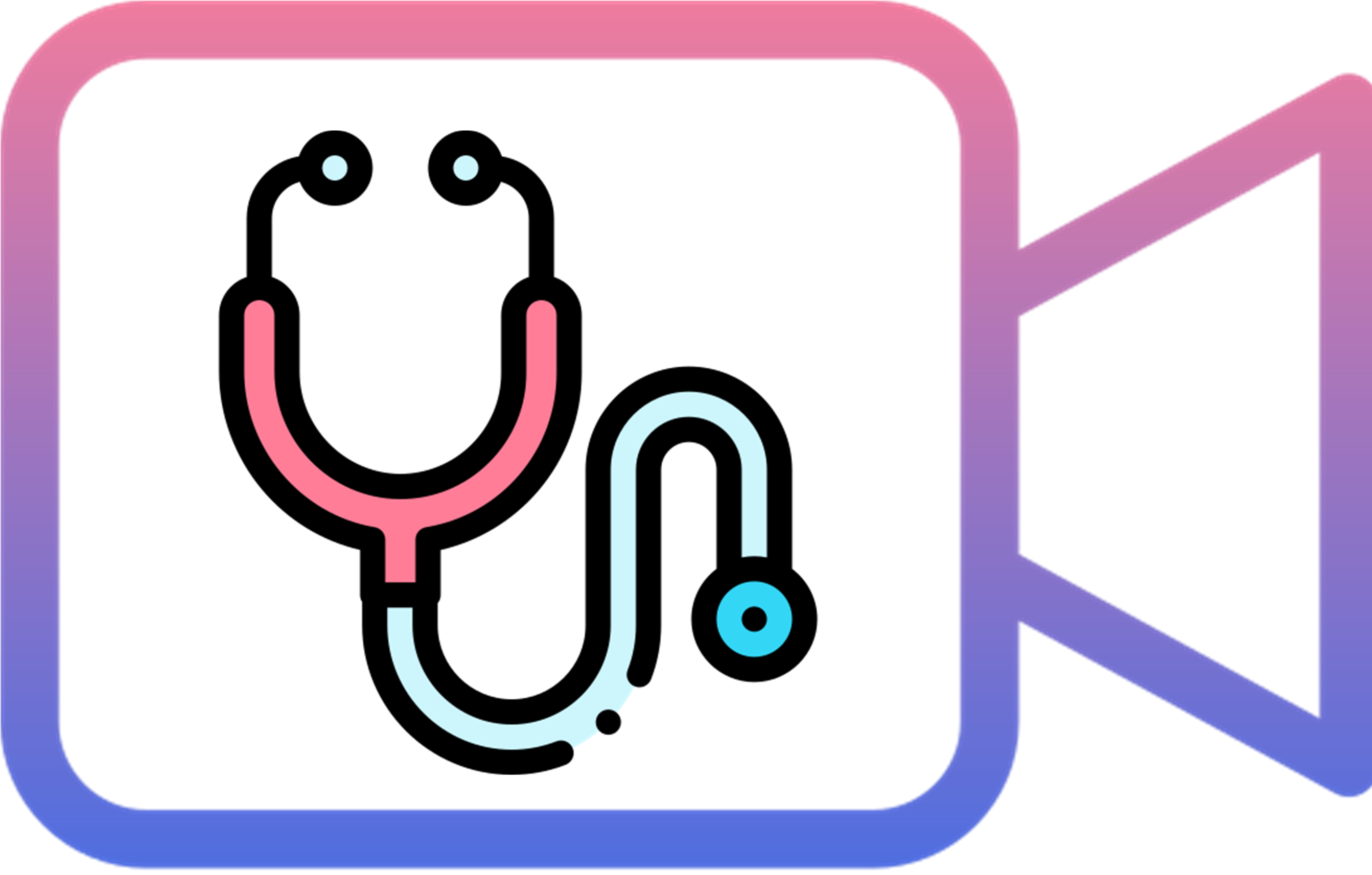}}\xspace}
\newcommand{\systemALogo}{\raisebox{-3pt}{\includegraphics[width=1.6em]{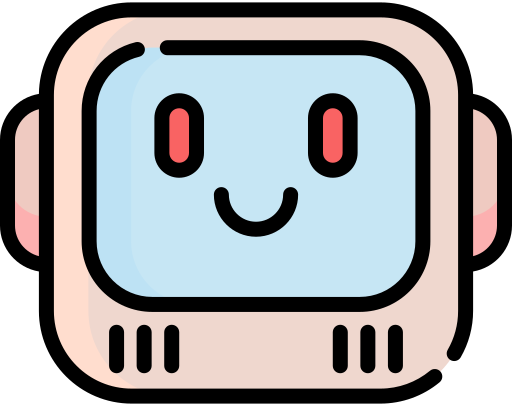}}\xspace}
\newcommand{\systemBLogo}{\raisebox{-3pt}{\includegraphics[width=1.7em]{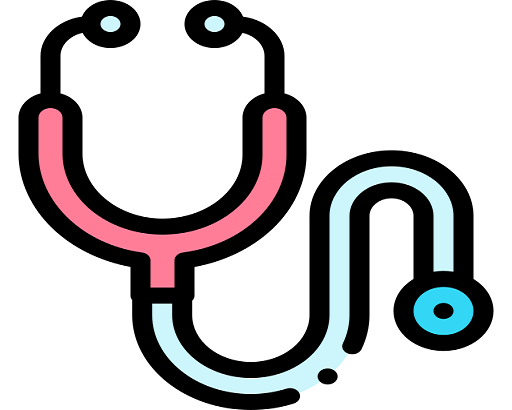}}\xspace}

\usepackage[table]{xcolor}
\usepackage{array}
\usepackage[most]{tcolorbox}

\definecolor{iccvblue}{rgb}{0.21,0.49,0.74}
\usepackage[pagebackref,breaklinks,colorlinks,allcolors=iccvblue]{hyperref}

\begin{document}


\title{\DrVLogo\texttt{Dr.V} : A Hierarchical Perception-Temporal-Cognition Framework to Diagnose Video Hallucination by Fine-grained Spatial-Temporal Grounding}


\author{
Meng Luo$^{1}$ \quad
Shengqiong Wu$^{1}$ \quad
Liqiang Jing$^{2}$ \quad
Tianjie Ju$^{1}$ \quad
Li Zheng$^{3}$ \quad
Jinxiang Lai$^{4}$ \quad \\
Tianlong Wu$^{1}$ \quad
Xinya Du$^{2}$ \quad
Jian Li$^{5}$ \quad
Siyuan Yan$^{6}$ \quad
Jiebo Luo$^{7}$ \quad
William Yang Wang$^{8}$ \quad \\
Hao Fei$^{1,\textrm{\Letter}}$ \quad
Mong-Li Lee$^{1}$ \quad
Wynne Hsu$^{1}$ \\
\textnormal{$^1$NUS} \qquad
\textnormal{$^2$UTD} \qquad
\textnormal{$^3$WHU} \qquad
\textnormal{$^4$HKUST} \qquad
\textnormal{$^5$NJU} \qquad
\textnormal{$^6$Monash} \qquad
\textnormal{$^7$UR} \qquad
\textnormal{$^8$UCSB}
}

\twocolumn[{%
\renewcommand\twocolumn[1][]{#1}%
\maketitle
\vspace{-8mm}
\begin{center}
   \includegraphics[width=0.95\linewidth]{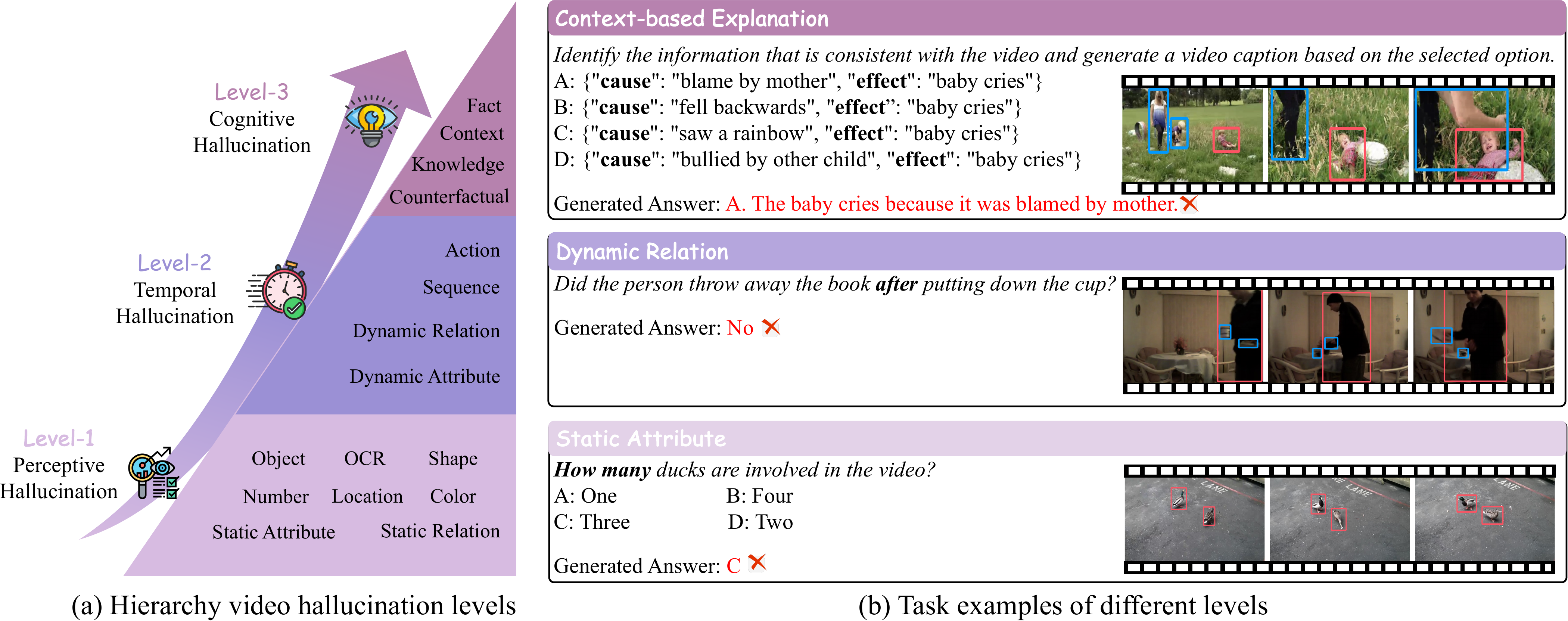}
    \vspace{-2mm}
    \captionof{figure}{(a) Hierarchical taxonomy of video hallucinations in LVMs, from Level-1 (perceptive) to Level-3 (cognitive), with increasing reasoning complexity. (b) Representative examples of hallucination types: static attribute, dynamic relation, and context-based explanation.}
\label{data-intro}
\end{center}%
}]

\def\thefootnote{\textrm{\Letter}}\footnotetext{Corresponding Author.}

\begin{abstract}
{Recent advancements in large video models (LVMs) have significantly enhance video understanding. However, these models continue to suffer from hallucinations, producing content that conflicts with input videos. To address this issue, we propose \textbf{\texttt{Dr.V}},
a hierarchical framework covering perceptive, temporal, and cognitive levels to diagnose video hallucination by fine-grained spatial-temporal grounding. Dr.V comprises of two key components: a benchmark dataset \textbf{\texttt{Dr.V-Bench}} and a satellite video agent \textbf{\texttt{Dr.V-Agent}}. Dr.V-Bench includes
10k instances drawn from 4,974 videos spanning diverse tasks, each enriched with detailed spatial-temporal annotation. 
Dr.V-Agent detects  hallucinations in LVMs by
systematically applying fine-grained spatial-temporal grounding at the perceptive and temporal levels, followed by cognitive level reasoning. This step-by-step pipeline mirrors human-like video comprehension and effectively identifies hallucinations. Extensive experiments demonstrate that Dr.V-Agent is effective in diagnosing hallucination while enhancing interpretability and reliability, offering a practical blueprint for robust video understanding in real-world scenarios.
All our data and code are available at \url{https://github.com/Eurekaleo/Dr.V}.} 
\end{abstract}

\vspace{-4mm}
\section{Introduction}
\label{sec:intro}

Video understanding has long been a crucial research area in artificial intelligence, focusing on the interpretation of dynamic content in videos, which is inherently more complex than static images. 
This complexity arises from the need to understand not only spatial visual semantics but also temporal sequences \cite{lavee2009understanding,zohar2024apollo,an2024mc,an2025unictokens}.
Built upon the foundation of large language models \cite{touvron2023llama,achiam2023gpt,bai2023qwen,guo2025deepseek}, large video models (LVMs) \cite{team2024gemini,chen2024internvl,bai2025qwen2,lin2025perceiveanythingrecognizeexplain} have significantly propelled progress in comprehending video content, such as video question answering \cite{lin2023video,zhang2024llavanext-video,feivideo}, video captioning \cite{xu2024pllava,cheng2024videollama,fei2024enhancing}. 
Despite these advancements, the generative nature of LVMs inevitably introduces a notable risk of ``hallucinations" – generating content that is inconsistent with the video content, misaligned with user intent, or factually incorrect \cite{liu2023models,bai2024hallucination,zhang2024eventhallusion}.
Such hallucinations undermine the reliability and trustworthiness of LVMs, posing critical challenges to their deployment in real-world scenarios.

To combat the hallucination issues in LVMs, it is essential to develop comprehensive and accurate evaluation protocols.
The research community has responded to this need by proposing several video hallucination benchmarks. 
Initial efforts, such as the pioneering VideoHallucer \cite{wang2024videohallucer}, established a foundational framework by categorizing hallucinations into intrinsic and extrinsic types.
Building on this, subsequent works have expanded the scope, often focusing on more specific facets of hallucination. 
For instance, some benchmarks target fine-grained temporal reasoning, including action sequences and scene transitions \cite{li2024vidhalluc, zhang2024eventhallusion}, while others concentrate on motion-related fallacies \cite{KongZCLYZ25} or action-scene inconsistencies \cite{bae2025mash}.
Other novel approaches have explored using synthetic videos to test for violations of common sense and physical laws \cite{li2025videohallu}, or have structured their evaluation along multiple dimensions like hallucination causes and question formats \cite{gao2025exploring}.

However, despite these valuable contributions, we argue that the current landscape of video hallucination evaluation suffers from two critical limitations, which prevent a truly holistic assessment:
\begin{compactitem}
\item \textbf{Fragmented and Incomplete Taxonomies.} Existing benchmarks, while strong in their respective niches, tend to address isolated aspects of hallucination. This specialization has led to a fragmented understanding of the problem, with no single benchmark offering a sufficiently comprehensive and unified taxonomy. Many common yet complex hallucination types, such as those involving dynamic attributes or nuanced context-based reasoning, remain underexplored. The coarse-grained or overly specialized categories used to date fail to capture the multi-dimensional nature of video hallucinations.
\item \textbf{Lack of Granularity in Annotation and Analysis.} The majority of current benchmarks are limited to instance-level labels, which can identify that a hallucination occurred, but not why or where. They often lack fine-grained annotations, such as the precise spatial-temporal grounding of the error within the video. This methodological gap makes it challenging to conduct root-cause analysis and develop targeted mitigation strategies, as the model's failure points remain unclear.
\end{compactitem}

To address the limitations of existing datasets, we introduce \textbf{\texttt{Dr.V-Bench}}, a novel benchmark designed to comprehensively evaluate video hallucinations.
First, we conduct an in-depth analysis of all possible types of video hallucinations and propose a hierarchical taxonomy based on the complexity of reasoning and levels of abstraction. As illustrated in Figure~\ref{data-intro}, the taxonomy comprises three levels. \textbf{Level 1: Perceptive Hallucination}, which includes errors in basic perceptive understanding such as \textit{object recognition}, \textit{numerical estimation}, \textit{color identification}, \textit{object localization}, \textit{static spatial relations}, and \textit{optical character recognition (OCR)}.
\textbf{Level 2: Temporal Hallucination}, which involves misinterpretations of temporal dynamics, including \textit{action recognition}, \textit{dynamic attribute recognition} (e.g., speed, motion direction), \textit{dynamic relational inference}, and \textit{event sequence understanding}.
\textbf{Level 3: Cognitive Hallucination}, which encompasses higher-level reasoning failures such as \textit{factual prediction}, \textit{counterfactual prediction}, \textit{context-based explanation}, and \textit{knowledge-based explanation}.
Second, we curate a large-scale collection of videos from diverse sources, emphasizing content that is complex, multi-faceted, and rich in temporal and semantic variation.
Third, we carefully design three evaluation tasks, i.e., \textit{Yes/No Question Answering (QA)}, \textit{Multiple-Choice QA}, and \textit{Video Captioning}, each tailored to probe the \textbf{14} identified hallucination types across different reasoning levels.
Moreover, we provide fine-grained annotations for each video instance, including detailed spatial-temporal grounding of objects relevant to the associated queries.
Finally, through a rigorous multi-stage human verification process, we construct a high-quality dataset containing \textbf{10,000} annotated instances paired with \textbf{4,974} videos. 
Each instance is enriched with spatial-temporal grounding information, enabling comprehensive evaluation of hallucination phenomena in LVMs.

Building upon the proposed benchmark, we further aim to develop a model capable of diagnosing and mitigating video hallucinations in LVMs.
Drawing inspiration from the mechanisms underlying human video comprehension, we revisit the root causes of video hallucination and present two key insights:
\textbf{First}, humans are capable of finely perceiving and localizing relevant content in a video, effortlessly identifying the target objects and the specific spatial-temporal segments corresponding to a given query. In contrast, LVMs often suffer from hallucinations due to their limited fine-grained spatial-temporal grounding capabilities. That is, their inability to accurately interpret pixel-level spatial details and maintain coherent temporal understanding over long video sequences.
\textbf{Second}, human reasoning typically unfolds in a structured, step-by-step manner: beginning with low-level perception, progressing through temporal inference, and culminating in high-level cognitive reasoning. However, existing LVMs often lack this staged reasoning process, leading to conflated or misaligned representations across different levels of abstraction.

\begin{figure*}[!t]
\centering
\includegraphics[width=0.9\textwidth]{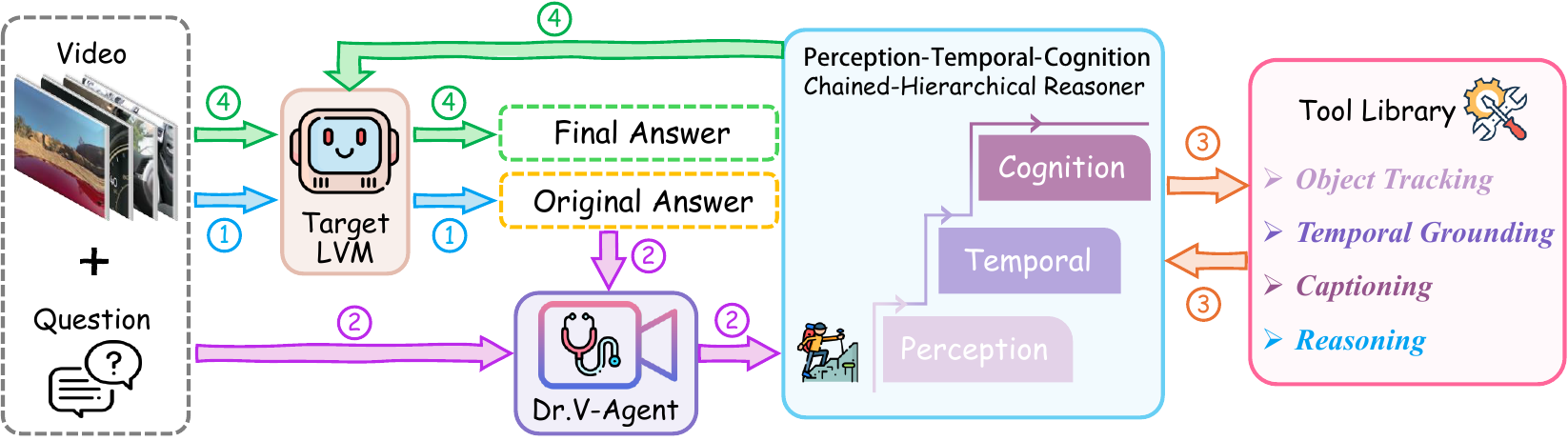}
\vspace{-2mm}
\caption{The scenario consists a target LVM and our \textbf{\texttt{Dr.V-Agent}} hallucination diagnosis system.
The overall workflow contains four steps:
{\color{steponecircle}\textcircled{1}} The target LVM generates an answer for a given video and QA input. 
{\color{steptwocircle}\textcircled{2}} Dr.V-Agent diagnoses hallucinations by analyzing the VQA pair and LVM's response using a chained-hierarchical reasoner. 
{\color{stepthreecircle}\textcircled{3}} The reasoner invokes expert tools to extract relevant video information for verification. 
{\color{stepfourcircle}\textcircled{4}} Dr.V-Agent generates feedback based on the analysis, prompting the target LVM to refine its response.}
\label{fig:method-intro}
\end{figure*}

Motivated by aforementioned observations, we propose a novel diagnostic model,\textbf{ \texttt{Dr.V-Agent}}, which emulates this hierarchical reasoning paradigm to identify and analyze hallucinations in LVMs. 
As shown in Figure \ref{fig:method-intro}, and inspired by the proposed hierarchical hallucination taxonomy, the core of \textbf{\texttt{Dr.V-Agent}} lies in a hierarchical chained reasoning mechanism, ``\textbf{From-Perception-to-Temporal-to-Cognition}'', which systematically diagnoses hallucination in a progressive, interpretable manner. 
Given an input question, video, and generated answers by an LVM, \textbf{\texttt{Dr.V-Agent}} initiates a progressive diagnostic process. 
At each stage of the reasoning chain, the agent selectively invokes advanced external tools to perform fine-grained spatial-temporal grounding directly on the raw video content, localizing relevant visual entities, tracking their temporal evolution, and verifying factual or contextual consistency with the question. 
Throughout this reasoning cascade, \textbf{\texttt{Dr.V-Agent}} continuously refines its understanding and, ultimately, produces a detailed diagnostic report that highlights the specific sources and types of hallucination present in the original LVM response. 
Based on this analysis, the LVM is guided to revise its answer, yielding a hallucination-free response.
We conduct extensive evaluations of \textbf{\texttt{Dr.V-Agent}} on the proposed benchmark, demonstrating its effectiveness in significantly reducing hallucination errors across a wide range of state-of-the-art (SOTA) LVMs. 
In addition, the interpretability and modular design of the agent provide transparent insights into model behavior and failure modes.

In summary, this paper makes the following key contributions:
\begin{compactitem}
    \item We first introduce the hierarchical hallucination taxonomy of video hallucinations that across three distinct reasoning levels: \textit{perception}, \textit{temporal}, and \textit{cognition}. This taxonomy provides a principled framework for analyzing hallucination phenomena in LVMs.
    
    \item We construct \texttt{Dr.V-Bench}, the most comprehensive and large-scale benchmark, which comprises 10,000 instances spanning 14 hallucination types across a wide range of complex video scenarios, and is enriched with fine-grained spatial-temporal grounding annotations to support detailed diagnostic analysis.
    
    \item We propose \texttt{Dr.V-Agent}, a novel diagnostic framework for video hallucination detection and mitigation. The agent performs a chained hierarchical reasoning progression and leverages advanced external tools for evidence verification, producing structured diagnostic feedback to guide LVMs toward generating more faithful and interpretable outputs. 

    \item We conduct extensive experiments to demonstrate the efficacy of the \texttt{Dr.V-Bench}, and effectiveness of the proposed \texttt{Dr.V-Agent} across multiple LVMs.
\end{compactitem}

\begin{figure*}[!t]
 \centering
 \includegraphics[width=0.9\linewidth]{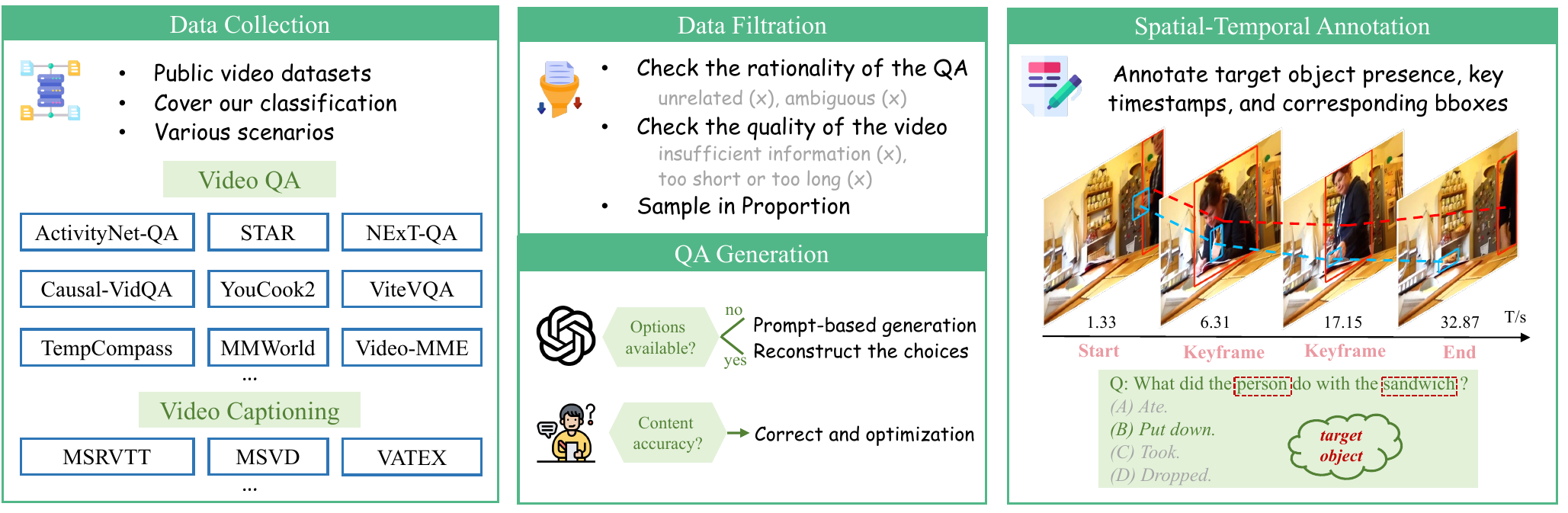}
 \vspace{-2mm}
 \caption{Data construction and annotation pipeline of our \textbf{\texttt{Dr.V-Bench}} dataset.}
 \label{Dr.V-Bench-annotation}
\end{figure*}

\vspace{-1mm}
\section{Related Work}
\label{sec:related_work}
\vspace{-1mm}

\subsection{LVMs and Hallucination}  
In recent years, the emergence of LLMs \cite{li2024survey,luo2024llm,fei2025path} has demonstrated unprecedented intelligence across various AI tasks. 
Building on top of LLMs, visual/video LLMs~\cite{wu2024towards,lin2023video,jin2024video,wu24next} (aka LVMs\footnote{`V' can refer to either `Visual' or `Video'; in this work, we focus on video LVMs.}) have also garnered significant research attention, leading to evident performance improvements in vision and video tasks. 
However, due to their generative nature, all these models inevitably produce outputs that deviate from the given input or reality, known as hallucination problems.
To combat hallucinations in LVMs, certain methods have been proposed. For example, Vista-LLaMA \cite{ma2024vista} optimizes the consistent distance between visual and language tokens. \citet{sun2024hallucination} propose optimizing video keyframe retrieval to address incorrect references in videos. \citet{zhang2024eventhallusion} introduce a Temporal Contrastive Decoding strategy to reduce event-related hallucinations in existing LVMs. \citet{wang2024videohallucer} attempt to prompt LVMs with self-reflection to mitigate hallucinations. Another approach, MASH-VLM \cite{bae2025mash}, specifically targets action-scene hallucination by disentangling spatial and temporal representations within the model, preventing the model from incorrectly inferring actions based on scene context, or vice versa.
More recently, methods based on Direct Preference Optimization (DPO)~\cite{rafailov2023direct} have become a prominent research direction for mitigating hallucinations.
These methods align model outputs with human or AI-provided preferences. 
For instance, some work leverages rewards from powerful language models to guide the DPO process, using detailed video captions as a proxy for video content to assess the factuality of generated responses \cite{zhang2024direct}.
Building on this, PaMi-VDPO \cite{ding2025pami} introduces an online preference learning framework that generates negative samples through video augmentation, guided by prompt awareness to avoid false rejections and better enforce video-response alignment.
Further pushing the boundaries of fine-grained alignment, VistaDPO \cite{huang2025vistadpo} proposes a hierarchical DPO framework that optimizes text-video preferences at three distinct levels: instance, temporal, and perceptive. 
Similarly, the work on HAVEN \cite{gao2025exploring} employs a combination of supervised reasoning fine-tuning and DPO to enhance the model's reasoning process and reduce hallucinations.
A viewpoint extensively verified in relevant studies \cite{sahoo2024comprehensive,bai2024hallucination} emphasizes that the root cause of visual hallucination lies in the lack of faithfulness to the given input video or facts, which can be attributed to the absence of fine-grained spatial-temporal-aware grounding. 
Only when models can accurately capture spatial-temporal details can they correctly perceive the existence, attributes, and relationships of objects in videos, understand the inherent dynamic temporal information, and further perform accurate semantic reasoning. 
Therefore, this work incorporates spatial-temporal grounding as a core component in the design of our video hallucination diagnosis system.

\subsection{Video Hallucination Benchmarks}
On the other hand, efforts should be directed towards the construction of video hallucination benchmarks, which are also key to addressing hallucination issues. 
Several related benchmarks have been proposed in recent years.
\citet{wang2024videohallucer} introduce the VideoHallucer, where video hallucinations are categorized into intrinsic and extrinsic types, resulting in a dataset of 1.8k Video QA instances. VidHalluc \cite{li2024vidhalluc} provides a dataset with 5k videos and defines three categories of hallucinations: action, temporal sequence, and scene transition. EventHallusion \cite{zhang2024eventhallusion} further focuses on mitigating video hallucinations by evaluating the event/action understanding of models. More recent benchmarks have expanded the scope and complexity.
For example, \citet{li2025videohallu} introduce VideoHallu, a benchmark focused on synthetic videos designed to test models' ability to detect violations of common sense and physical laws. \citet{gao2025exploring} present HAVEN, a comprehensive benchmark with 6k questions built upon three dimensions: hallucination causes, hallucination aspects, and question formats. 
The MASH-VLM work is accompanied by the UNSCENE benchmark, which is specifically curated to evaluate action-scene hallucination by using videos with unusual contexts or scene-only content \cite{bae2025mash}.
Furthermore, to support preference-based learning methods, VistaDPO provides 7.2k question-answer pairs annotated with both chosen and rejected responses \cite{huang2025vistadpo}.
We observe that none of these existing video hallucination benchmarks provide a comprehensive taxonomy definition for video hallucinations, which should be a critical aspect. 
Through an extensive survey, we have identified that all sources of video hallucination can be categorized into three major dimensions: \emph{Perception}, \emph{Temporal}, and \emph{Cognition}, each of which can be further subdivided into more specific subcategories. 
Moreover, we include high-quality spatial-temporal grounding annotations as an essential part of our benchmark, resulting in the largest-scale video hallucination dataset to date, with 10k instances, the most comprehensive category definitions, and the most diverse video scenes in the community.

\begin{figure*}[!t]
\centering
\includegraphics[width=0.9\linewidth]{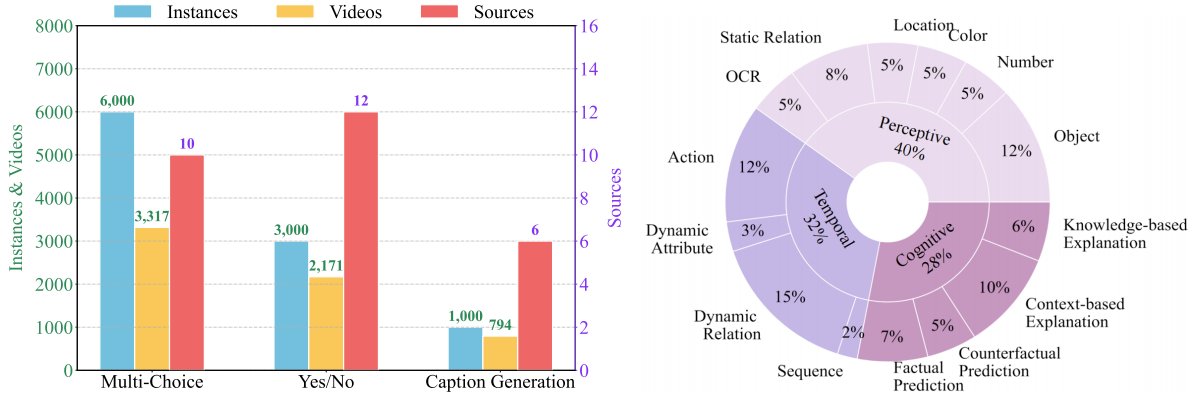}
\vspace{-2mm}
\caption{Main statistics of our \textbf{\texttt{Dr.V-Bench}} dataset. The left bar chart presents the number of instances, videos, and dataset sources for different task types, and the right pie chart illustrates the proportion of each hallucination type.}
\label{fig:method-intro-stats}
\end{figure*}

\vspace{-1mm}
\section{Preliminaries}\label{sec2}
\vspace{-1mm}
Video hallucination refers to a phenomenon in which an LVM may produce hallucinated output that conflicts with factual content or even invent entirely new objects, scenes, or actions to the input video.
Existing LVMs typically exhibit hallucinations across various video-related tasks. 

\paratitle{Hallucination-diagnosis System.} Given a target LVM \systemALogo, capable of video comprehension, that takes as input a video \(V\) and a textual prompt \({T}\) corresponding to a task query 
and outputs an answer \({A}\) after reasoning.
Our goal is to develop a hallucination-diagnosis system \systemBLogo 
that is independent of the target LVM to diagnose potential hallucinations by analyzing the answer $A$ from the target LVM based on the raw input video \(V\) and \({T}\).
If \({A}\) contains a hallucination, our system \systemBLogo will generate a rationale pinpointing the specifics of the hallucination, which serves as feedback to the target \systemALogo to refine its response.

\paratitle{Taxonomy of Video Hallucination.} 
Video hallucinations differ significantly from those in static images due to the rich spatial semantics and complex temporal dynamics inherent in videos. 
These additional layers of complexity mean that video hallucinations can manifest across a spectrum of levels, from low-level perceptive and temporal aspects to high-order cognitive functions. 
As such, we propose a new taxonomy of video hallucination. 
Inspired by human intuition, we categorize hallucinations hierarchically into 3 levels, covering a total of 14 fine-grained types.

\vspace{-1mm}
\section{{Dr.V-Bench}: A Comprehensive Benchmark for Video Hallucination}
\vspace{-1mm}

We introduce \texttt{Dr.V-Bench}, a novel and comprehensive benchmark meticulously designed to evaluate and diagnose hallucinations in LVMs. This section details its construction pipeline, quality control protocols, statistical properties, and advancements over existing benchmarks. Figure~\ref{Dr.V-Bench-annotation} provides a high-level overview of the entire data construction and annotation process.

\vspace{-1mm}
\subsection{Data Collection and Curation}
\vspace{-1mm}
The foundation of \texttt{Dr.V-Bench} is built upon a diverse collection of 15 well-established, public video datasets. 
These source datasets provide a vast pool of existing QA pairs, predominantly in a multiple-choice format, as shown in Table~\ref{qa_examples}. This allows us to ground our work in questions that are already validated by the community. We don't create QA from scratch; instead, we meticulously curate and transform existing data for the specific purpose of hallucination analysis.
Before this transformation, we perform a rigorous manual curation process on the original data. We review the original QA pairs to filter out any that are ambiguous, factually incorrect, or unrelated to the video content. We also assess video quality, excluding videos that are of unreasonable length or lack sufficient information for a definitive answer. This ensures that the data foundation for our benchmark is clean and reliable.

\begin{figure*}[!t]
 \centering
 \includegraphics[width=0.9\linewidth]{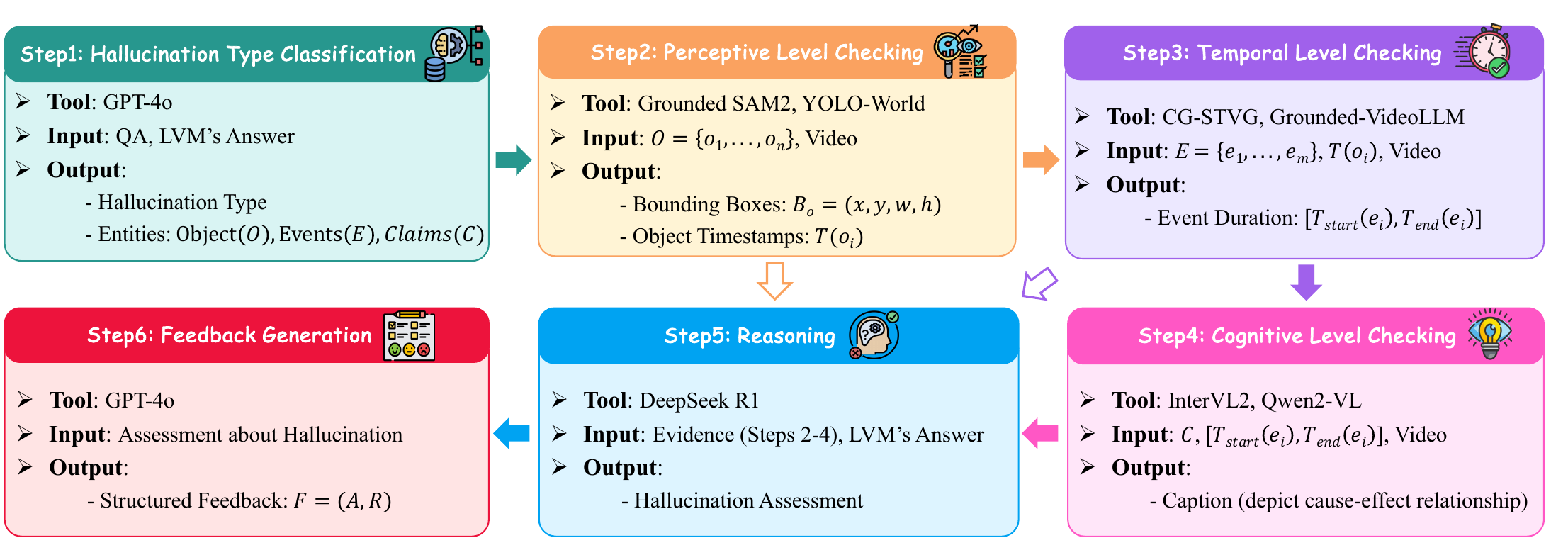}
 \vspace{-2mm}
 \caption{Our \textbf{\texttt{Dr.V-Agent}} framework diagnoses and locates hallucinations through a chained perception-temporal-cognition reasoning process. Different hallucinations will select different reasoning paths.}
 \label{fig:Dr.V-Agent}
 \vspace{-1mm}
\end{figure*}

\subsection{Hallucination-centric QA Generation}
Following curation, we restructure the validated QA pairs to facilitate the diagnosis of hallucinations through three key steps: taxonomic classification, reconstruction of answer options, and format diversification.
First, we meticulously map each curated QA pair to our proposed taxonomy of hallucinations. 
Each question is analyzed and categorized into one of the fine-grained hallucination types under the three levels.
This step imposes a systematic, hierarchical structure on the previously uncategorized data, enabling targeted evaluation of specific hallucination vulnerabilities.
Second, we reconstruct the answer options to create challenging, hallucination-centric test cases. The original incorrect options in the source datasets are often trivial or not designed to mimic plausible model failures. To address this, we design specific prompt templates (see Appendix $\S$\ref{sec:data}) tailored to each hallucination type. 
These templates guide GPT-4o \cite{hurst2024gpt} to generate new incorrect answers that are deliberately misleading and act as convincing hallucinated foils to the ground-truth answer. This reconstruction transforms a standard QA task into a rigorous test of a model's ability to resist specific types of hallucinations.
Finally, based on the restructured multiple-choice QA, we diversify the task formats to create a multi-faceted evaluation suite. 
We design three distinct QA formats—yes/no, multiple-choice, and caption generation—to assess both discriminative and generative model capabilities.

\subsection{Fine-grained Spatial-Temporal Grounding}
To enable fine-grained analysis, we manually annotate the videos with precise spatial-temporal information. The annotation process consists of four steps: extraction of target objects, annotation of start and end frames, annotation of key frames, and annotation of bounding boxes. Detailed annotation procedures are provided in Appendix~\S\ref{sec:label}.

\subsection{Dataset Statistics}
\texttt{Dr.V-Bench} is a large-scale benchmark that sets a new standard in the field. It contains 10k instances distributed across 4.9k unique videos. The dataset covers 50 diverse scenarios/domains, such as daily scenes, life recordings, artistic performances, and sports competitions. The tasks are broken down into 3k yes/no QA, 6k multiple-choice QA, and 1k caption generation QA instances, as illustrated in Appendix $\S$\ref{sec:data}.
Notably, around 25\% of the videos feature significant complexity (e.g., multiple scene transitions, concurrent overlapped events, and questions requiring higher-level reasoning), ensuring a rigorous test of an LVM's capabilities. This comprehensive design ensures sufficient generalizability for LVMs to adapt to tasks and contents requiring domain-specific reasoning. An overview of the dataset statistics is provided in Figure~\ref{fig:method-intro-stats}.

\vspace{-1mm}
\section{{Dr.V-Agent} for Mitigating Hallucination}
\vspace{-1mm}
\texttt{Dr.V-Agent} is an agentic system that leverages LLMs to dynamically integrate various SOTA external tools. It employs a chained-hierarchical perception-temporal-cognition reasoning process to diagnose and reason about hallucinations in a fine-grained manner.
The framework dynamically selects a reasoning path based on the potential hallucination's complexity, as illustrated in Figure \ref{fig:Dr.V-Agent}. 
For instance, perceptive level hallucinations are diagnosed via an abbreviated path (Steps 1, 2, 5, and 6), and temporal level issues require the inclusion of temporal checking (Steps 1, 2, 3, 5, and 6), while only the cognitive level hallucinations necessitate the complete six-step reasoning chain. This hierarchical approach not only mirrors the logical dependencies from perception to cognition but also ensures an efficient and targeted analysis.

\paratitle{Step 1: Hallucination Type Classification.}  
In this step, we use GPT-4o to analyze the QA pair and the target LVM’s response to extract 
the set of objects denoted as $O = \{o_1, \dots, o_n\}$, events $E = \{e_1, \dots, e_m\}$, and causal claims $C = \{c_1, \dots, c_k\}$ where $C$ denotes cause-and-effect relationships stated in the QA.

\paratitle{Step 2: Perceptive Level Checking.}
Perceptive level understanding requires accurate object identification and relative spatial positioning. 
To ensure high robustness and mitigate the risk of errors from any single tool, we employ a dual-tool approach for cross-validation. 
Specifically, we use Grounded SAM 2~\cite{ren2024grounded} andYOLO-World~\cite{Cheng2024YOLOWorld} for open-set object detection and grounding. 
For each object $o \in O$, we enhance detection precision by taking the intersection of the bounding boxes predicted by both tools, and its timestamp is determined by averaging the timestamps they report.

\begin{table*}[t!]
\fontsize{8}{10}\selectfont 
\setlength{\tabcolsep}{1.12mm}
\centering
\vspace{-3mm}
\begin{tabular}{@{}lccccccccccccccc@{}}
\toprule
\multirow{2}{*}{\textbf{Methods}} & \multicolumn{6}{c}{\textbf{Perceptive}} & \multicolumn{4}{c}{\textbf{Temporal}} & \multicolumn{4}{c}{\textbf{Cognitive}} & \multirow{2}{*}{\textbf{Avg}} \\ 
\cmidrule(r){2-7} \cmidrule(r){8-11} \cmidrule(r){12-15}
 & \textbf{Obj.} & \textbf{Col.} & \textbf{Num.} & \textbf{Loc.} & \textbf{SRel.} & \textbf{OCR} & \textbf{Act.} & \textbf{Atr.} & \textbf{DRel.} & \textbf{Seq.} & \textbf{Fct.} & \textbf{CnFct.} & \textbf{Cxt.} & \textbf{Knk.}  \\ 
\midrule
\rowcolor[gray]{0.9} \multicolumn{16}{@{}l}{\textit{\textbf{Open-source LVMs}}} \\
VideoChat2 (7B)~\cite{li2024mvbench} & 41.88 & 38.15 & \underline{30.14} & 64.22 & 36.20 & \underline{30.50} & \underline{34.06} & \underline{28.61} & \underline{37.77} & \underline{28.52} & 36.68 & \underline{33.96} & \underline{36.35} & \underline{30.85} & \underline{36.28} \\
Video-ChatGPT (7B)~\cite{maaz2023video} & \underline{39.05} & \underline{35.15} & 36.30 & \underline{60.62} & \underline{33.00} & 41.68 & 35.09 & 34.94 & 38.19 & 30.78 & \underline{33.61} & 36.36 & 38.28 & 39.05 & 38.01 \\
Video-LLaVA (7B)~\cite{lin2023video} & 53.57 & 50.58 & 48.98 & 80.93 & 56.65 & 42.86 & 38.38 & 39.93 & 45.09 & 45.02 & 48.49 & 39.31 & 61.16 & 50.03 & 50.07 \\
LLaMA-VID (7B)~\cite{li2025llama} & 52.42 & 44.64 & 46.52 & 78.72 & 55.94 & 51.40 & 39.08 & 40.84 & 43.05 & 37.99 & 45.99 & 51.66 & 61.03 & 53.86 & 50.22 \\
LLaVA-NeXT-Video-DPO (7B)~\cite{zhang2024llavanext-video} & 54.76 & 58.32 & 58.02 & 83.65 & 63.39 & 49.25 & 48.19 & 44.79 & 49.47 & 39.00 & 45.76 & 60.77 & 70.64 & 57.46 &  56.80\\
PLLaVA (7B)~\cite{xu2024pllava} & 68.00 & 65.80 & 73.03 & 83.56 & 76.65 & 57.54 & 43.60 & 49.00 & 55.93 & 44.50 & 59.71 & 65.66 & 71.53 & 61.73 &  62.58\\
InternVL2 (8B)~\cite{chen2024internvl} & 73.88 & 72.95 & 72.82 & 90.93 & 76.14 & 64.20 & 60.66 & 50.96 & 64.84 & 52.83 & 63.47 & 61.22 & 74.55 & 64.38 &  67.42\\
Qwen2-VL (7B)~\cite{wang2024qwen2} & 76.57 & 76.66 & 72.84 & 90.33 & 79.31 & 73.55 & 65.72 & 55.81 & 70.55 & 64.02 & 70.01 & 68.83 & 81.21 & 71.91 &  72.67 \\
\midrule
\rowcolor[gray]{0.9} \multicolumn{16}{@{}l}{\textit{\textbf{Closed-source LVMs}}} \\
GPT-4o~\cite{hurst2024gpt} & 81.25 & 81.16 & 78.03 & 93.02 & 82.81 & 78.51 & 72.58 & 60.83 & 75.20 & 68.76 & 74.33 & 73.74 & 86.03 & 75.85 & 77.29 \\
Gemini-1.5-Pro~\cite{team2024gemini} & \textbf{83.29} & \textbf{83.55} & \textbf{80.90} & \textbf{94.04} & \textbf{84.97} & \textbf{80.34} & \textbf{75.76} & \textbf{63.51} & \textbf{77.76} & \textbf{70.54} & \textbf{77.20} & \textbf{76.46} & \textbf{88.24} & \textbf{79.01} & \textbf{79.68} \\
\midrule
Human & 98.54 & 99.63 & 95.80 & 99.50 & 99.00 & 98.05 & 90.12 & 89.22 & 96.50 & 89.59 & 96.00 & 90.40 & 99.50 & 91.62 & 95.25 \\  
\bottomrule
\end{tabular}
\caption{Performance of LVMs on \textbf{\texttt{Dr.V-Bench}}. Columns represent the hallucination types: Object (Obj.), Color (Col.), Number (Num.), Location (Loc.), Static Relation (SRel.), OCR; Action (Act.), Dynamic Attribute (Atr.), Dynamic relation (DRel.), Sequence (Seq.); Factual Prediction (Fct.), Counterfactual Prediction (CnFct.), Context-based Explanation (Cxt.), Knowledge-based Explanation (Knk.).}
\label{tab:types}
\vspace{-2mm}
\end{table*}

\paratitle{Step 3: Temporal Level Checking.}
This step is activated only for temporal and cognitive level inquiries. 
Temporal understanding requires event detection and verifying temporal order. 
Following the same principle of robust, cross-validated analysis, we utilize CG-STVG~\cite{gu2024context} and Grounded-VideoLLM~\cite{wang2024grounded} for temporal video grounding. 
With the events identified and cross-verified by these models, we then verify whether the events occur and whether their temporal order aligns with the expected sequence based on the ground truth.

\paratitle{Step 4: Cognitive Level Checking.}
Cognitive level understanding requires integrating spatial-temporal cues with commonsense or contextual knowledge to interpret causal relationships. 
For each causal claim in $C$, we use InternVL2~\cite{chen2024internvl} and Qwen2-VL~\cite{bai2025qwen2} to generate dense descriptive captions that explicitly depict cause-and-effect relationships. The captioning process focuses on the corresponding event duration detected in Step 3.

\paratitle{Step 5: Reasoning.}  
We diagnose hallucinations in the target LVM’s response by utilizing DeepSeek R1~\cite{guo2025deepseek} to identify inconsistencies between the spatial, temporal, and cognitive information obtained in Steps 2–4 and the responses generated by the LVM.

\paratitle{Step 6: Feedback Generation.}  
This final step consolidates the detected hallucinations into a structured format, providing clear feedback for the LVM to improve its understanding capability.  
For each detected hallucination, we generate the feedback  $\mathcal{F} = (\mathcal{A}, \mathcal{R})$
where \( \mathcal{A} \) is the extracted spatial-temporal-causal information, including objects \(\mathcal{O}\) with their bounding boxes, events \(\mathcal{E}\) with their timestamps, and causal claims \(\mathcal{C}\), as well as supporting captions, and 
\( \mathcal{R} \) consists of suggestions for LVM to refine its response.

Our design of \texttt{Dr.V-Agent} offers three key advantages over foundational video models. First, \texttt{Dr.V-Agent} has the flexibility to utilize SOTA tools, thereby overcoming bottlenecks in the target LVM that may cause hallucinations. For example, GPT-4o serves as the core LLM for planning and reasoning, while models such as SAM2 are used for precise spatial-temporal grounding in videos. Second, our solution operates under a training-free paradigm, eliminating the need for retraining or fine-tuning on additional datasets. Third, its adaptive, hierarchical reasoning process ensures efficiency by tailoring the computational path to the complexity of the hallucination, thus avoiding the overhead associated with a fixed, monolithic pipeline.

\vspace{-1mm}
\section{Performance and Discussions}
\vspace{-1mm}
\paratitle{Settings.}
To comprehensively evaluate the capabilities of LVMs on our VideoQA tasks, we select ten representative models: eight open-source LVMs (VideoChat2~\cite{li2024mvbench}, Video-ChatGPT~\cite{maaz2023video}, Video-LLaVA~\cite{lin2023video}, LLaMA-VID~\cite{li2025llama}, LLaVA-NeXT-Video-DPO~\cite{zhang2024llavanext-video}, PLLaVA~\cite{xu2024pllava}, Qwen2-VL, and InternVL2) and two advanced closed-source models (GPT-4o and Gemini-1.5-Pro~\cite{team2024gemini}). For a fair comparison, all models are evaluated using their default hyper-parameters, including ``max\_new\_tokens", ``do\_sample", ``temperature", and ``num\_of\_frames". For closed-source models, we set the frame sampling rate to 1 and limit the maximum number of frames to 128 to ensure evaluation efficiency while maintaining sufficient temporal coverage.

\begin{table*}[!t]
\centering
\renewcommand{\arraystretch}{1.1}
\resizebox{\linewidth}{!}{
\begin{tabular}{lccccccccccccccc}
\toprule
\textbf{Model} & \textbf{Obj.} & \textbf{Col.} & \textbf{Num.} & \textbf{Loc.} & \textbf{SRel.} & \textbf{OCR} & \textbf{Act.} & \textbf{Atr.} & \textbf{DRel.} & \textbf{Seq.} & \textbf{Fct.} & \textbf{CnFct.} & \textbf{Cxt.} & \textbf{Knk.} & \textbf{Avg.}\\ 
\midrule
VideoChat2  & 41.88 & 38.15 & 30.14 & 64.22 & 36.20 & 30.50 & 34.06 & 28.61 & 37.77 & 28.52 & 36.68 & 33.96 & 36.35 & 30.85 & \textbf{38.01} \\
\quad {+ Self-PEP} & 49.52 & 43.60 & 35.17 & 66.81 & 45.12 & 39.10 & 37.19 & 31.04 & 45.52 & 32.70 & 43.65 & 37.22 & 43.91 & 34.68 & 44.82 (\textcolor{red}{+6.81}) \\
\quad {+ Dr.V-Agent} & 65.40 & 54.91 & 45.62 & 72.18 & 63.61 & 56.95 & 43.69 & 36.09 & 61.55 & 41.38 & 58.13 & 43.99 & 59.61 & 42.62 & \textbf{53.43} (\textcolor{red}{+15.42}) \\
\midrule
LLaVA-NeXT  & 54.76 & 58.32 & 58.02 & 83.65 & 63.39 & 49.25 & 48.19 & 44.79 & 49.47 & 39.00 & 45.76 & 60.77 & 70.64 & 57.46 & \textbf{56.80} \\
\quad {+ Self-PEP} & 49.24 & 54.39 & 54.40 & 81.79 & 56.97 & 43.04 & 45.93 & 43.04 & 43.90 & 35.99 & 40.74 & 58.42 & 65.20 & 54.70 & 52.45 (\textcolor{red}{-4.35}) \\
\quad {+ Dr.V-Agent} & 76.76 & 74.01 & 72.49 & 91.09 & 89.04 & 74.01 & 57.20 & 51.79 & 71.72 & 51.02 & 65.82 & 70.15 & 92.39 & 68.48 & \textbf{74.21} (\textcolor{red}{+17.41}) \\
\midrule
Qwen2-VL  & 76.57 & 76.66 & 72.84 & 90.33 & 79.31 & 73.55 & 65.72 & 55.81 & 70.55 & 64.02 & 70.01 & 68.83 & 81.21 & 71.91 & \textbf{72.67} \\
\quad {+ Self-PEP} & 80.37 & 79.37 & 75.34 & 91.61 & 83.73 & 77.83 & 67.27 & 57.02 & 74.39 & 66.10 & 73.47 & 70.45 & 84.96 & 73.81 & 75.67 (\textcolor{red}{+3.00}) \\
\quad {+ Dr.V-Agent} & 89.19 & 85.66 & 81.14 & 94.60 & 94.03 & 87.76 & 70.89 & 59.83 & 83.33 & 70.92 & 81.53 & 74.21 & 93.69 & 78.23 & \textbf{82.64} (\textcolor{red}{+9.97}) \\
\midrule
GPT-4o  & 81.25 & 81.16 & 78.03 & 93.02 & 82.81 & 78.51 & 72.58 & 60.83 & 75.20 & 68.76 & 74.33 & 73.74 & 86.03 & 75.85 & \textbf{77.29} \\
\quad {+ Self-PEP} & 87.63 & 85.70 & 82.22 & 95.18 & 90.23 & 85.69 & 75.19 & 62.86 & 81.65 & 72.24 & 80.14 & 76.46 & 92.33 & 79.04 & 82.33 (\textcolor{red}{+5.04}) \\
\quad {+ Dr.V-Agent} & 95.23 & 91.13 & 87.23 & 97.75 & 98.12 & 94.26 & 78.31 & 66.28 & 89.36 & 76.40 & 87.30 & 79.71 & 99.06 & 82.90 & \textbf{88.36} (\textcolor{red}{+11.07}) \\
\bottomrule
\end{tabular}
}
\vspace{-2mm}
\caption{Representative results comparing Self-PEP vs. \textbf{\texttt{Dr.V-Agent}} on four indicative LVMs.}
\label{improvement_main}
\end{table*}

\begin{figure}[!t]
 \centering
 \includegraphics[width=0.93\linewidth]{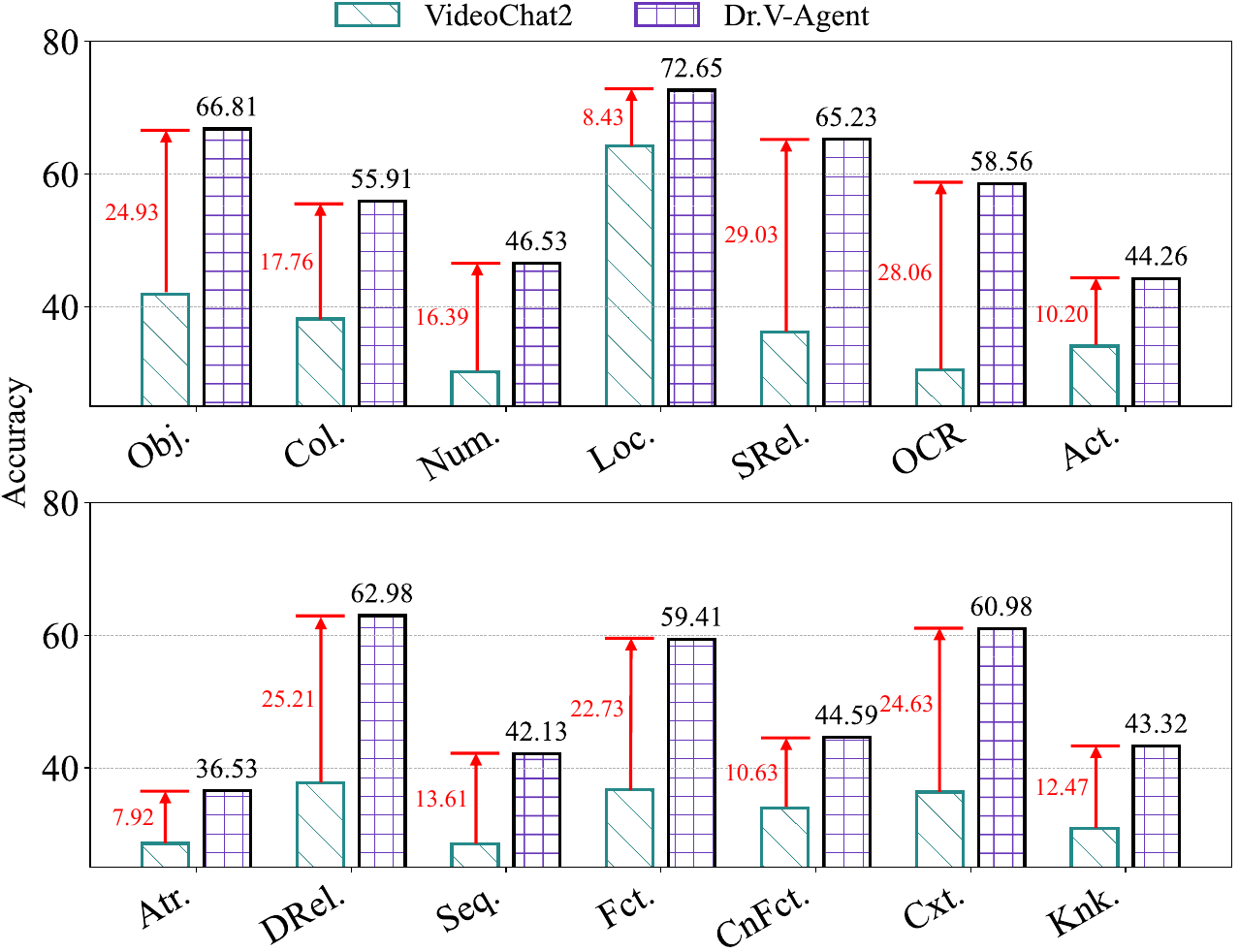}
 \vspace{-2mm}
 \caption{Performance gains across hallucination types for VideoChat2 when equipped with \textbf{\texttt{Dr.V-Agent}}.}
 \label{fig:typeperformance}
\vspace{-2mm}
\end{figure}

\vspace{-1mm}
\subsection{Performance of LVMs on Dr.V-Bench}
We evaluate a range of open-source and closed-source LVMs on \texttt{Dr.V-Bench}. The comprehensive results are presented in Table~\ref{tab:types}. Overall, \texttt{Dr.V-Bench} proves to be a highly challenging benchmark, with all tested models exhibiting significant hallucinations. 
A clear performance hierarchy emerges from the results. Models generally perform best on perceptive tasks, with accuracy declining substantially for temporal and cognitive tasks. This indicates that while current LVMs have developed a reasonable capacity for static scene understanding, they lack the more rigorous spatial-temporal understanding and advanced reasoning required for complex video analysis. For instance, even the top-performing open-source model, Qwen2-VL, sees its accuracy drop from 78.75\% on perceptive tasks to 65.61\% on temporal tasks in the multiple-choice QA setting.
We also observe a considerable performance gap between model families. Closed-source models like Gemini-1.5-Pro and GPT-4o consistently outperform open-source models across all categories, particularly in temporal and cognitive reasoning. While recent open-source models like Qwen2-VL and InternVL2 have narrowed this gap, a significant disparity remains, especially in challenging temporal tasks like ``Sequence'' understanding and cognitive tasks like ``Counterfactual Prediction'', as shown in Table~\ref{tab:types}. Nonetheless, no model yet approaches human-level proficiency, highlighting that video hallucination is a severe and unsolved problem.

\subsection{Evaluating the Effectiveness of Dr.V-Agent}

To validate the effectiveness of our proposed \texttt{Dr.V-Agent}, we conduct a comparative analysis against a strong and relevant baseline. We chose Self-PEP \cite{wang2024videohallucer}, a representative and recent self-correction strategy, as it shares our goal of mitigating hallucinations in a plug-and-play manner. 

\paratitle{Quantitative Comparison.}
As shown in Table~\ref{improvement_main}, equipping LVMs with \texttt{Dr.V-Agent} consistently and substantially outperforms using the Self-PEP strategy across representative models and hallucination types. While Self-PEP yields moderate improvements in some cases, its effectiveness is inconsistent and can even lead to performance degradation compared to the vanilla model (e.g., -4.35\% on LLaVA-NeXT; -5.30\% on PLLaVA). In contrast, \texttt{Dr.V-Agent} achieves significant and robust gains across all tested models, with particularly pronounced improvement for lower-performing LVMs like VideoChat2 (+18.60\%). Complete results for all models and hallucination types are provided in Appendix Table~\ref{improvement_full}.

\paratitle{Analysis of Performance and Efficiency.}
The superiority of \texttt{Dr.V-Agent} stems from two key advantages: its performance architecture and its efficiency.
On Performance, unlike Self-PEP's reliance on internal knowledge which can be flawed, \texttt{Dr.V-Agent}'s success comes from its ability to leverage external, specialized tools for verification. By integrating fine-grained and reliable video information through these tools, it overcomes the limitations of self-correction and allows the base LVM to make more informed decisions. As illustrated in Figure~\ref{fig:typeperformance}, this leads to substantial improvements across specific hallucination types like ``Obj.'', ``SRel.'', ``OCR'', and ``DRel.'', confirming that grounding the model's reasoning in precise, externally-verified spatial-temporal information is pivotal for mitigating hallucinations.
On Efficiency, \texttt{Dr.V-Agent} introduces a paradigm shift. Its most critical advantage is being training-free. Unlike monolithic LVMs that require costly pre-training and fine-tuning, our agent operates by intelligently composing existing expert tools. This eliminates prohibitive computational overhead and enhances accessibility. Furthermore, its modular design makes it inherently flexible and future-proof. As better tools for object detection or temporal grounding emerge, they can be seamlessly integrated into framework without retraining the entire system, ensuring sustained high performance with minimal effort.

\vspace{-1mm}
\section{Conclusion}
\vspace{-1mm}
We propose \textbf{\texttt{Dr.V}}, a hierarchical framework to address hallucinations in LVMs. The \textbf{\texttt{Dr.V-Bench}} benchmark establishes a three-tiered evaluation protocol—perceptive, temporal, and cognitive levels—supported by 10k instances with spatial-temporal annotations across diverse video scenarios. Our \textbf{\texttt{Dr.V-Agent}} framework systematically mitigates hallucinations through a structured pipeline: it first verifies fine-grained spatial-temporal grounding, then performs cognitive-level reasoning to align outputs to video content. Experiments demonstrate that this approach significantly reduces hallucination rates while enhancing interpretability, offering a practical and robust solution for reliable video understanding in real-world applications.

{
    \small
    \bibliographystyle{ieeenat_fullname}
    \bibliography{main}

\begin{thebibliography}{68}
\providecommand{\natexlab}[1]{#1}
\providecommand{\url}[1]{\texttt{#1}}
\expandafter\ifx\csname urlstyle\endcsname\relax
  \providecommand{\doi}[1]{doi: #1}\else
  \providecommand{\doi}{doi: \begingroup \urlstyle{rm}\Url}\fi

\bibitem[Achiam et~al.(2023)Achiam, Adler, Agarwal, Ahmad, Akkaya, Aleman, Almeida, Altenschmidt, Altman, Anadkat, et~al.]{achiam2023gpt}
Josh Achiam, Steven Adler, Sandhini Agarwal, Lama Ahmad, Ilge Akkaya, Florencia~Leoni Aleman, Diogo Almeida, Janko Altenschmidt, Sam Altman, Shyamal Anadkat, et~al.
\newblock Gpt-4 technical report.
\newblock \emph{arXiv preprint arXiv:2303.08774}, 2023.

\bibitem[An et~al.(2024)An, Yang, Lu, Zhang, Zeng, Luo, Cao, Liang, Chen, She, et~al.]{an2024mc}
Ruichuan An, Sihan Yang, Ming Lu, Renrui Zhang, Kai Zeng, Yulin Luo, Jiajun Cao, Hao Liang, Ying Chen, Qi She, et~al.
\newblock Mc-llava: Multi-concept personalized vision-language model.
\newblock \emph{arXiv preprint arXiv:2411.11706}, 2024.

\bibitem[An et~al.(2025)An, Yang, Zhang, Shen, Lu, Dai, Liang, Guo, Yan, Luo, et~al.]{an2025unictokens}
Ruichuan An, Sihan Yang, Renrui Zhang, Zijun Shen, Ming Lu, Gaole Dai, Hao Liang, Ziyu Guo, Shilin Yan, Yulin Luo, et~al.
\newblock Unictokens: Boosting personalized understanding and generation via unified concept tokens.
\newblock \emph{arXiv preprint arXiv:2505.14671}, 2025.

\bibitem[Bae et~al.(2025)Bae, Kim, Lee, Lee, Lee, and Choi]{bae2025mash}
Kyungho Bae, Jinhyung Kim, Sihaeng Lee, Soonyoung Lee, Gunhee Lee, and Jinwoo Choi.
\newblock Mash-vlm: Mitigating action-scene hallucination in video-llms through disentangled spatial-temporal representations.
\newblock In \emph{Proceedings of the Computer Vision and Pattern Recognition Conference}, pages 13744--13753, 2025.

\bibitem[Bai et~al.(2023)Bai, Bai, Chu, Cui, Dang, Deng, Fan, Ge, Han, Huang, et~al.]{bai2023qwen}
Jinze Bai, Shuai Bai, Yunfei Chu, Zeyu Cui, Kai Dang, Xiaodong Deng, Yang Fan, Wenbin Ge, Yu Han, Fei Huang, et~al.
\newblock Qwen technical report.
\newblock \emph{arXiv preprint arXiv:2309.16609}, 2023.

\bibitem[Bai et~al.(2025)Bai, Chen, Liu, Wang, Ge, Song, Dang, Wang, Wang, Tang, et~al.]{bai2025qwen2}
Shuai Bai, Keqin Chen, Xuejing Liu, Jialin Wang, Wenbin Ge, Sibo Song, Kai Dang, Peng Wang, Shijie Wang, Jun Tang, et~al.
\newblock Qwen2. 5-vl technical report.
\newblock \emph{arXiv preprint arXiv:2502.13923}, 2025.

\bibitem[Bai et~al.(2024)Bai, Wang, Xiao, He, Han, Zhang, and Shou]{bai2024hallucination}
Zechen Bai, Pichao Wang, Tianjun Xiao, Tong He, Zongbo Han, Zheng Zhang, and Mike~Zheng Shou.
\newblock Hallucination of multimodal large language models: A survey.
\newblock \emph{arXiv preprint arXiv:2404.18930}, 2024.

\bibitem[Chen and Dolan(2011)]{chen2011collecting}
David Chen and William~B Dolan.
\newblock Collecting highly parallel data for paraphrase evaluation.
\newblock In \emph{Proceedings of the 49th annual meeting of the association for computational linguistics: human language technologies}, pages 190--200, 2011.

\bibitem[Chen et~al.(2024)Chen, Wu, Wang, Su, Chen, Xing, Zhong, Zhang, Zhu, Lu, et~al.]{chen2024internvl}
Zhe Chen, Jiannan Wu, Wenhai Wang, Weijie Su, Guo Chen, Sen Xing, Muyan Zhong, Qinglong Zhang, Xizhou Zhu, Lewei Lu, et~al.
\newblock Internvl: Scaling up vision foundation models and aligning for generic visual-linguistic tasks.
\newblock In \emph{Proceedings of the IEEE/CVF Conference on Computer Vision and Pattern Recognition}, pages 24185--24198, 2024.

\bibitem[Cheng et~al.(2024{\natexlab{a}})Cheng, Song, Ge, Liu, Wang, and Shan]{Cheng2024YOLOWorld}
Tianheng Cheng, Lin Song, Yixiao Ge, Wenyu Liu, Xinggang Wang, and Ying Shan.
\newblock Yolo-world: Real-time open-vocabulary object detection.
\newblock In \emph{Proc. IEEE Conf. Computer Vision and Pattern Recognition (CVPR)}, 2024{\natexlab{a}}.

\bibitem[Cheng et~al.(2024{\natexlab{b}})Cheng, Leng, Zhang, Xin, Li, Chen, Zhu, Zhang, Luo, Zhao, et~al.]{cheng2024videollama}
Zesen Cheng, Sicong Leng, Hang Zhang, Yifei Xin, Xin Li, Guanzheng Chen, Yongxin Zhu, Wenqi Zhang, Ziyang Luo, Deli Zhao, et~al.
\newblock Videollama 2: Advancing spatial-temporal modeling and audio understanding in video-llms.
\newblock \emph{arXiv preprint arXiv:2406.07476}, 2024{\natexlab{b}}.

\bibitem[Choong et~al.(2024)Choong, Guo, and Kankanhalli]{choong2024vidhal}
Wey~Yeh Choong, Yangyang Guo, and Mohan Kankanhalli.
\newblock Vidhal: Benchmarking temporal hallucinations in vision llms.
\newblock \emph{arXiv preprint arXiv:2411.16771}, 2024.

\bibitem[Chu et~al.(2024)Chu, Zhang, Sun, Xue, Wang, Qin, and Ren]{chu2024sora}
Zhixuan Chu, Lei Zhang, Yichen Sun, Siqiao Xue, Zhibo Wang, Zhan Qin, and Kui Ren.
\newblock Sora detector: A unified hallucination detection for large text-to-video models.
\newblock \emph{arXiv preprint arXiv:2405.04180}, 2024.

\bibitem[Ding et~al.(2025)Ding, Zhang, Han, Hong, Xu, and Li]{ding2025pami}
Xinpeng Ding, Kui Zhang, Jianhua Han, Lanqing Hong, Hang Xu, and Xiaomeng Li.
\newblock Pami-vdpo: Mitigating video hallucinations by prompt-aware multi-instance video preference learning.
\newblock \emph{arXiv preprint arXiv:2504.05810}, 2025.

\bibitem[Fei et~al.()Fei, Wu, Ji, Zhang, Zhang, Lee, and Hsu]{feivideo}
Hao Fei, Shengqiong Wu, Wei Ji, Hanwang Zhang, Meishan Zhang, Mong-Li Lee, and Wynne Hsu.
\newblock Video-of-thought: Step-by-step video reasoning from perception to cognition.
\newblock In \emph{Forty-first International Conference on Machine Learning}.

\bibitem[Fei et~al.(2024)Fei, Wu, Zhang, Zhang, Chua, and Yan]{fei2024enhancing}
Hao Fei, Shengqiong Wu, Meishan Zhang, Min Zhang, Tat-Seng Chua, and Shuicheng Yan.
\newblock Enhancing video-language representations with structural spatio-temporal alignment.
\newblock \emph{IEEE Transactions on Pattern Analysis and Machine Intelligence}, 2024.

\bibitem[Fei et~al.(2025)Fei, Zhou, Li, Li, Xu, Li, Wu, Wang, Zhou, Meng, et~al.]{fei2025path}
Hao Fei, Yuan Zhou, Juncheng Li, Xiangtai Li, Qingshan Xu, Bobo Li, Shengqiong Wu, Yaoting Wang, Junbao Zhou, Jiahao Meng, et~al.
\newblock On path to multimodal generalist: General-level and general-bench.
\newblock In \emph{Proceedings of the International Conference on Machine Learning}, 2025.

\bibitem[Fu et~al.(2024)Fu, Dai, Luo, Li, Ren, Zhang, Wang, Zhou, Shen, Zhang, et~al.]{fu2024video}
Chaoyou Fu, Yuhan Dai, Yongdong Luo, Lei Li, Shuhuai Ren, Renrui Zhang, Zihan Wang, Chenyu Zhou, Yunhang Shen, Mengdan Zhang, et~al.
\newblock Video-mme: The first-ever comprehensive evaluation benchmark of multi-modal llms in video analysis.
\newblock \emph{arXiv preprint arXiv:2405.21075}, 2024.

\bibitem[Gao et~al.(2025)Gao, Qu, Tang, Bi, Liu, Chen, Liang, Su, and Huang]{gao2025exploring}
Hongcheng Gao, Jiashu Qu, Jingyi Tang, Baolong Bi, Yue Liu, Hongyu Chen, Li Liang, Li Su, and Qingming Huang.
\newblock Exploring hallucination of large multimodal models in video understanding: Benchmark, analysis and mitigation.
\newblock \emph{arXiv preprint arXiv:2503.19622}, 2025.

\bibitem[Gu et~al.(2024)Gu, Fan, Huang, Luo, and Zhang]{gu2024context}
Xin Gu, Heng Fan, Yan Huang, Tiejian Luo, and Libo Zhang.
\newblock Context-guided spatio-temporal video grounding.
\newblock In \emph{Proceedings of the IEEE/CVF Conference on Computer Vision and Pattern Recognition}, pages 18330--18339, 2024.

\bibitem[Guo et~al.(2025)Guo, Yang, Zhang, Song, Zhang, Xu, Zhu, Ma, Wang, Bi, et~al.]{guo2025deepseek}
Daya Guo, Dejian Yang, Haowei Zhang, Junxiao Song, Ruoyu Zhang, Runxin Xu, Qihao Zhu, Shirong Ma, Peiyi Wang, Xiao Bi, et~al.
\newblock Deepseek-r1: Incentivizing reasoning capability in llms via reinforcement learning.
\newblock \emph{arXiv preprint arXiv:2501.12948}, 2025.

\bibitem[He et~al.(2024)He, Feng, Zheng, Lu, Zhu, Li, Fan, Wang, Li, Yang, et~al.]{he2024mmworld}
Xuehai He, Weixi Feng, Kaizhi Zheng, Yujie Lu, Wanrong Zhu, Jiachen Li, Yue Fan, Jianfeng Wang, Linjie Li, Zhengyuan Yang, et~al.
\newblock Mmworld: Towards multi-discipline multi-faceted world model evaluation in videos.
\newblock \emph{arXiv preprint arXiv:2406.08407}, 2024.

\bibitem[Huang et~al.(2025)Huang, Chen, Wu, Luo, Fu, Du, Zhang, and Fei]{huang2025vistadpo}
Haojian Huang, Haodong Chen, Shengqiong Wu, Meng Luo, Jinlan Fu, Xinya Du, Hanwang Zhang, and Hao Fei.
\newblock Vistadpo: Video hierarchical spatial-temporal direct preference optimization for large video models.
\newblock \emph{arXiv preprint arXiv:2504.13122}, 2025.

\bibitem[Hurst et~al.(2024)Hurst, Lerer, Goucher, Perelman, Ramesh, Clark, Ostrow, Welihinda, Hayes, Radford, et~al.]{hurst2024gpt}
Aaron Hurst, Adam Lerer, Adam~P Goucher, Adam Perelman, Aditya Ramesh, Aidan Clark, AJ Ostrow, Akila Welihinda, Alan Hayes, Alec Radford, et~al.
\newblock Gpt-4o system card.
\newblock \emph{arXiv preprint arXiv:2410.21276}, 2024.

\bibitem[Jin et~al.(2024)Jin, Sun, Xu, Chen, Jiang, Huang, Song, Liu, Zhang, Song, et~al.]{jin2024video}
Yang Jin, Zhicheng Sun, Kun Xu, Liwei Chen, Hao Jiang, Quzhe Huang, Chengru Song, Yuliang Liu, Di Zhang, Yang Song, et~al.
\newblock Video-lavit: Unified video-language pre-training with decoupled visual-motional tokenization.
\newblock In \emph{International Conference on Machine Learning}, pages 22185--22209, 2024.

\bibitem[Kong et~al.(2025)Kong, Zeng, Chen, Li, Yan, and Zhu]{KongZCLYZ25}
Ming Kong, Xianzhou Zeng, Luyuan Chen, Yadong Li, Bo Yan, and Qiang Zhu.
\newblock Mhbench: Demystifying motion hallucination in videollms.
\newblock In \emph{AAAI-25, Sponsored by the Association for the Advancement of Artificial Intelligence, February 25 - March 4, 2025, Philadelphia, PA, {USA}}, pages 4401--4409, 2025.

\bibitem[Lavee et~al.(2009)Lavee, Rivlin, and Rudzsky]{lavee2009understanding}
Gal Lavee, Ehud Rivlin, and Michael Rudzsky.
\newblock Understanding video events: A survey of methods for automatic interpretation of semantic occurrences in video.
\newblock \emph{IEEE Transactions on Systems, Man, and Cybernetics, Part C (Applications and Reviews)}, 2009.

\bibitem[Li et~al.(2024{\natexlab{a}})Li, Im, and Fazli]{li2024vidhalluc}
Chaoyu Li, Eun~Woo Im, and Pooyan Fazli.
\newblock Vidhalluc: Evaluating temporal hallucinations in multimodal large language models for video understanding.
\newblock \emph{arXiv preprint arXiv:2412.03735}, 2024{\natexlab{a}}.

\bibitem[Li et~al.(2022)Li, Niu, and Zhang]{li2022representation}
Jiangtong Li, Li Niu, and Liqing Zhang.
\newblock From representation to reasoning: Towards both evidence and commonsense reasoning for video question-answering.
\newblock In \emph{Proceedings of the IEEE/CVF conference on computer vision and pattern recognition}, pages 21273--21282, 2022.

\bibitem[Li et~al.(2024{\natexlab{b}})Li, Lu, Fei, Luo, Dai, Xia, Jin, Gan, Qi, Fu, et~al.]{li2024survey}
Jian Li, Weiheng Lu, Hao Fei, Meng Luo, Ming Dai, Min Xia, Yizhang Jin, Zhenye Gan, Ding Qi, Chaoyou Fu, et~al.
\newblock A survey on benchmarks of multimodal large language models.
\newblock \emph{arXiv preprint arXiv:2408.08632}, 2024{\natexlab{b}}.

\bibitem[Li et~al.(2024{\natexlab{c}})Li, Wang, He, Li, Wang, Liu, Wang, Xu, Chen, Luo, et~al.]{li2024mvbench}
Kunchang Li, Yali Wang, Yinan He, Yizhuo Li, Yi Wang, Yi Liu, Zun Wang, Jilan Xu, Guo Chen, Ping Luo, et~al.
\newblock Mvbench: A comprehensive multi-modal video understanding benchmark.
\newblock In \emph{Proceedings of the IEEE/CVF Conference on Computer Vision and Pattern Recognition}, pages 22195--22206, 2024{\natexlab{c}}.

\bibitem[Li et~al.(2025{\natexlab{a}})Li, Wang, and Jia]{li2025llama}
Yanwei Li, Chengyao Wang, and Jiaya Jia.
\newblock Llama-vid: An image is worth 2 tokens in large language models.
\newblock In \emph{European Conference on Computer Vision}, pages 323--340, 2025{\natexlab{a}}.

\bibitem[Li et~al.(2025{\natexlab{b}})Li, Wu, Shi, Qin, Du, Zhou, Manocha, and Boyd-Graber]{li2025videohallu}
Zongxia Li, Xiyang Wu, Guangyao Shi, Yubin Qin, Hongyang Du, Tianyi Zhou, Dinesh Manocha, and Jordan~Lee Boyd-Graber.
\newblock Videohallu: Evaluating and mitigating multi-modal hallucinations on synthetic video understanding.
\newblock \emph{arXiv preprint arXiv:2505.01481}, 2025{\natexlab{b}}.

\bibitem[Lin et~al.(2023)Lin, Ye, Zhu, Cui, Ning, Jin, and Yuan]{lin2023video}
Bin Lin, Yang Ye, Bin Zhu, Jiaxi Cui, Munan Ning, Peng Jin, and Li Yuan.
\newblock Video-llava: Learning united visual representation by alignment before projection.
\newblock \emph{arXiv preprint arXiv:2311.10122}, 2023.

\bibitem[Lin et~al.(2025)Lin, Wei, An, Ren, Chen, Zhang, Guo, Zhang, Zhang, and Li]{lin2025perceiveanythingrecognizeexplain}
Weifeng Lin, Xinyu Wei, Ruichuan An, Tianhe Ren, Tingwei Chen, Renrui Zhang, Ziyu Guo, Wentao Zhang, Lei Zhang, and Hongsheng Li.
\newblock Perceive anything: Recognize, explain, caption, and segment anything in images and videos, 2025.

\bibitem[Liu and Wan(2023)]{liu2023models}
Hui Liu and Xiaojun Wan.
\newblock Models see hallucinations: Evaluating the factuality in video captioning.
\newblock \emph{arXiv preprint arXiv:2303.02961}, 2023.

\bibitem[Liu et~al.(2023)Liu, Zeng, Ren, Li, Zhang, Yang, Li, Yang, Su, Zhu, et~al.]{liu2023grounding}
Shilong Liu, Zhaoyang Zeng, Tianhe Ren, Feng Li, Hao Zhang, Jie Yang, Chunyuan Li, Jianwei Yang, Hang Su, Jun Zhu, et~al.
\newblock Grounding dino: Marrying dino with grounded pre-training for open-set object detection.
\newblock \emph{arXiv preprint arXiv:2303.05499}, 2023.

\bibitem[Liu et~al.(2024)Liu, Li, Liu, Wang, Ren, Li, Chen, Sun, and Hou]{liu2024tempcompass}
Yuanxin Liu, Shicheng Li, Yi Liu, Yuxiang Wang, Shuhuai Ren, Lei Li, Sishuo Chen, Xu Sun, and Lu Hou.
\newblock Tempcompass: Do video llms really understand videos?
\newblock \emph{arXiv preprint arXiv:2403.00476}, 2024.

\bibitem[Luo et~al.(2024)Luo, An, Zou, Tang, Liu, and Zhang]{luo2024llm}
Yulin Luo, Ruichuan An, Bocheng Zou, Yiming Tang, Jiaming Liu, and Shanghang Zhang.
\newblock Llm as dataset analyst: Subpopulation structure discovery with large language model.
\newblock In \emph{European Conference on Computer Vision}, pages 235--252. Springer, 2024.

\bibitem[Ma et~al.(2024)Ma, Jin, Wang, Xian, Feng, and Yang]{ma2024vista}
Fan Ma, Xiaojie Jin, Heng Wang, Yuchen Xian, Jiashi Feng, and Yi Yang.
\newblock Vista-llama: Reducing hallucination in video language models via equal distance to visual tokens.
\newblock In \emph{Proceedings of the IEEE/CVF Conference on Computer Vision and Pattern Recognition}, pages 13151--13160, 2024.

\bibitem[Maaz et~al.(2023)Maaz, Rasheed, Khan, and Khan]{maaz2023video}
Muhammad Maaz, Hanoona Rasheed, Salman Khan, and Fahad~Shahbaz Khan.
\newblock Video-chatgpt: Towards detailed video understanding via large vision and language models.
\newblock \emph{arXiv preprint arXiv:2306.05424}, 2023.

\bibitem[Maaz et~al.(2024)Maaz, Rasheed, Khan, and Khan]{maaz2024videogpt}
Muhammad Maaz, Hanoona Rasheed, Salman Khan, and Fahad Khan.
\newblock Videogpt+: Integrating image and video encoders for enhanced video understanding.
\newblock \emph{arXiv preprint arXiv:2406.09418}, 2024.

\bibitem[Rafailov et~al.(2023)Rafailov, Sharma, Mitchell, Manning, Ermon, and Finn]{rafailov2023direct}
Rafael Rafailov, Archit Sharma, Eric Mitchell, Christopher~D Manning, Stefano Ermon, and Chelsea Finn.
\newblock Direct preference optimization: Your language model is secretly a reward model.
\newblock \emph{Advances in neural information processing systems}, 36:\penalty0 53728--53741, 2023.

\bibitem[Rawte et~al.(2024)Rawte, Jain, Sinha, Kaushik, Bansal, Vishwanath, Jain, Reganti, Jain, Chadha, et~al.]{rawte2024vibe}
Vipula Rawte, Sarthak Jain, Aarush Sinha, Garv Kaushik, Aman Bansal, Prathiksha~Rumale Vishwanath, Samyak~Rajesh Jain, Aishwarya~Naresh Reganti, Vinija Jain, Aman Chadha, et~al.
\newblock Vibe: A text-to-video benchmark for evaluating hallucination in large multimodal models.
\newblock \emph{arXiv preprint arXiv:2411.10867}, 2024.

\bibitem[Ren et~al.(2024)Ren, Liu, Zeng, Lin, Li, Cao, Chen, Huang, Chen, Yan, Zeng, Zhang, Li, Yang, Li, Jiang, and Zhang]{ren2024grounded}
Tianhe Ren, Shilong Liu, Ailing Zeng, Jing Lin, Kunchang Li, He Cao, Jiayu Chen, Xinyu Huang, Yukang Chen, Feng Yan, Zhaoyang Zeng, Hao Zhang, Feng Li, Jie Yang, Hongyang Li, Qing Jiang, and Lei Zhang.
\newblock Grounded sam: Assembling open-world models for diverse visual tasks, 2024.

\bibitem[Sahoo et~al.(2024)Sahoo, Meharia, Ghosh, Saha, Jain, and Chadha]{sahoo2024comprehensive}
Pranab Sahoo, Prabhash Meharia, Akash Ghosh, Sriparna Saha, Vinija Jain, and Aman Chadha.
\newblock A comprehensive survey of hallucination in large language, image, video and audio foundation models.
\newblock \emph{Findings of the Association for Computational Linguistics: EMNLP 2024}, pages 11709--11724, 2024.

\bibitem[Shangguan et~al.(2024)Shangguan, Li, Ding, Zheng, Zhao, Fitzgerald, and Cohan]{shangguan2024tomato}
Ziyao Shangguan, Chuhan Li, Yuxuan Ding, Yanan Zheng, Yilun Zhao, Tesca Fitzgerald, and Arman Cohan.
\newblock Tomato: Assessing visual temporal reasoning capabilities in multimodal foundation models.
\newblock \emph{arXiv preprint arXiv:2410.23266}, 2024.

\bibitem[Sun et~al.(2024)Sun, Liu, Liu, Pu, Zhang, and Xie]{sun2024hallucination}
Yiwei Sun, Zhihang Liu, Chuanbin Liu, Bowei Pu, Zhihan Zhang, and Hongtao Xie.
\newblock Hallucination mitigation prompts long-term video understanding.
\newblock \emph{arXiv preprint arXiv:2406.11333}, 2024.

\bibitem[Team et~al.(2024)Team, Georgiev, Lei, Burnell, Bai, Gulati, Tanzer, Vincent, Pan, Wang, et~al.]{team2024gemini}
Gemini Team, Petko Georgiev, Ving~Ian Lei, Ryan Burnell, Libin Bai, Anmol Gulati, Garrett Tanzer, Damien Vincent, Zhufeng Pan, Shibo Wang, et~al.
\newblock Gemini 1.5: Unlocking multimodal understanding across millions of tokens of context.
\newblock \emph{arXiv preprint arXiv:2403.05530}, 2024.

\bibitem[Tom et~al.(2023)Tom, Mathew, Garcia-Bordils, Karatzas, and Jawahar]{tom2023reading}
George Tom, Minesh Mathew, Sergi Garcia-Bordils, Dimosthenis Karatzas, and CV Jawahar.
\newblock Reading between the lanes: Text videoqa on the road.
\newblock In \emph{International Conference on Document Analysis and Recognition}, pages 137--154. Springer, 2023.

\bibitem[Touvron et~al.(2023)Touvron, Martin, Stone, Albert, Almahairi, Babaei, Bashlykov, Batra, Bhargava, Bhosale, et~al.]{touvron2023llama}
Hugo Touvron, Louis Martin, Kevin Stone, Peter Albert, Amjad Almahairi, Yasmine Babaei, Nikolay Bashlykov, Soumya Batra, Prajjwal Bhargava, Shruti Bhosale, et~al.
\newblock Llama 2: Open foundation and fine-tuned chat models.
\newblock \emph{arXiv preprint arXiv:2307.09288}, 2023.

\bibitem[Wang et~al.(2024{\natexlab{a}})Wang, Xu, Cheng, Diao, Zhou, Cao, Wang, Ge, and Huang]{wang2024grounded}
Haibo Wang, Zhiyang Xu, Yu Cheng, Shizhe Diao, Yufan Zhou, Yixin Cao, Qifan Wang, Weifeng Ge, and Lifu Huang.
\newblock Grounded-videollm: Sharpening fine-grained temporal grounding in video large language models.
\newblock \emph{arXiv preprint arXiv:2410.03290}, 2024{\natexlab{a}}.

\bibitem[Wang et~al.(2024{\natexlab{b}})Wang, Bai, Tan, Wang, Fan, Bai, Chen, Liu, Wang, Ge, et~al.]{wang2024qwen2}
Peng Wang, Shuai Bai, Sinan Tan, Shijie Wang, Zhihao Fan, Jinze Bai, Keqin Chen, Xuejing Liu, Jialin Wang, Wenbin Ge, et~al.
\newblock Qwen2-vl: Enhancing vision-language model's perception of the world at any resolution.
\newblock \emph{arXiv preprint arXiv:2409.12191}, 2024{\natexlab{b}}.

\bibitem[Wang et~al.(2019)Wang, Wu, Chen, Li, Wang, and Wang]{wang2019vatex}
Xin Wang, Jiawei Wu, Junkun Chen, Lei Li, Yuan-Fang Wang, and William~Yang Wang.
\newblock Vatex: A large-scale, high-quality multilingual dataset for video-and-language research.
\newblock In \emph{Proceedings of the IEEE/CVF international conference on computer vision}, pages 4581--4591, 2019.

\bibitem[Wang et~al.(2024{\natexlab{c}})Wang, Wang, Zhao, Xie, and Zheng]{wang2024videohallucer}
Yuxuan Wang, Yueqian Wang, Dongyan Zhao, Cihang Xie, and Zilong Zheng.
\newblock Videohallucer: Evaluating intrinsic and extrinsic hallucinations in large video-language models.
\newblock \emph{arXiv preprint arXiv:2406.16338}, 2024{\natexlab{c}}.

\bibitem[Wu et~al.(2024{\natexlab{a}})Wu, Fei, Li, Ji, Zhang, Chua, and Yan]{wu2024towards}
Shengqiong Wu, Hao Fei, Xiangtai Li, Jiayi Ji, Hanwang Zhang, Tat-Seng Chua, and Shuicheng Yan.
\newblock Towards semantic equivalence of tokenization in multimodal llm.
\newblock \emph{arXiv preprint arXiv:2406.05127}, 2024{\natexlab{a}}.

\bibitem[Wu et~al.(2024{\natexlab{b}})Wu, Fei, Qu, Ji, and Chua]{wu24next}
Shengqiong Wu, Hao Fei, Leigang Qu, Wei Ji, and Tat-Seng Chua.
\newblock {NE}x{T}-{GPT}: Any-to-any multimodal {LLM}.
\newblock In \emph{Proceedings of the International Conference on Machine Learning}, pages 53366--53397, 2024{\natexlab{b}}.

\bibitem[Xiao et~al.(2021)Xiao, Shang, Yao, and Chua]{xiao2021next}
Junbin Xiao, Xindi Shang, Angela Yao, and Tat-Seng Chua.
\newblock Next-qa: Next phase of question-answering to explaining temporal actions.
\newblock In \emph{Proceedings of the IEEE/CVF conference on computer vision and pattern recognition}, pages 9777--9786, 2021.

\bibitem[Xu et~al.(2016)Xu, Mei, Yao, and Rui]{xu2016msr}
Jun Xu, Tao Mei, Ting Yao, and Yong Rui.
\newblock Msr-vtt: A large video description dataset for bridging video and language.
\newblock In \emph{Proceedings of the IEEE conference on computer vision and pattern recognition}, pages 5288--5296, 2016.

\bibitem[Xu et~al.(2024)Xu, Zhao, Zhou, Lin, Ng, and Feng]{xu2024pllava}
Lin Xu, Yilin Zhao, Daquan Zhou, Zhijie Lin, See~Kiong Ng, and Jiashi Feng.
\newblock Pllava: Parameter-free llava extension from images to videos for video dense captioning.
\newblock \emph{arXiv preprint arXiv:2404.16994}, 2024.

\bibitem[Yi et~al.(2019)Yi, Gan, Li, Kohli, Wu, Torralba, and Tenenbaum]{yi2019clevrer}
Kexin Yi, Chuang Gan, Yunzhu Li, Pushmeet Kohli, Jiajun Wu, Antonio Torralba, and Joshua~B Tenenbaum.
\newblock Clevrer: Collision events for video representation and reasoning.
\newblock \emph{arXiv preprint arXiv:1910.01442}, 2019.

\bibitem[Yu et~al.(2019)Yu, Xu, Yu, Yu, Zhao, Zhuang, and Tao]{yu2019activitynet}
Zhou Yu, Dejing Xu, Jun Yu, Ting Yu, Zhou Zhao, Yueting Zhuang, and Dacheng Tao.
\newblock Activitynet-qa: A dataset for understanding complex web videos via question answering.
\newblock In \emph{Proceedings of the AAAI Conference on Artificial Intelligence}, 2019.

\bibitem[Zhang et~al.(2024{\natexlab{a}})Zhang, Jiao, Chen, Chen, and Jiang]{zhang2024eventhallusion}
Jiacheng Zhang, Yang Jiao, Shaoxiang Chen, Jingjing Chen, and Yu-Gang Jiang.
\newblock Eventhallusion: Diagnosing event hallucinations in video llms.
\newblock \emph{arXiv preprint arXiv:2409.16597}, 2024{\natexlab{a}}.

\bibitem[Zhang et~al.(2024{\natexlab{b}})Zhang, Gui, Sun, Feng, Xu, Zhang, Fu, Li, Hauptmann, Bisk, et~al.]{zhang2024direct}
Ruohong Zhang, Liangke Gui, Zhiqing Sun, Yihao Feng, Keyang Xu, Yuanhan Zhang, Di Fu, Chunyuan Li, Alexander Hauptmann, Yonatan Bisk, et~al.
\newblock Direct preference optimization of video large multimodal models from language model reward.
\newblock \emph{arXiv preprint arXiv:2404.01258}, 2024{\natexlab{b}}.

\bibitem[Zhang et~al.(2024{\natexlab{c}})Zhang, Li, Liu, Lee, Gui, Fu, Feng, Liu, and Li]{zhang2024llavanext-video}
Yuanhan Zhang, Bo Li, haotian Liu, Yong~jae Lee, Liangke Gui, Di Fu, Jiashi Feng, Ziwei Liu, and Chunyuan Li.
\newblock Llava-next: A strong zero-shot video understanding model, 2024{\natexlab{c}}.

\bibitem[Zhao et~al.(2022)Zhao, Li, Wang, Li, Zhou, Zhang, Xuyang, Yu, Yu, Li, et~al.]{zhao2022towards}
Minyi Zhao, Bingjia Li, Jie Wang, Wanqing Li, Wenjing Zhou, Lan Zhang, Shijie Xuyang, Zhihang Yu, Xinkun Yu, Guangze Li, et~al.
\newblock Towards video text visual question answering: Benchmark and baseline.
\newblock \emph{Advances in Neural Information Processing Systems}, 35:\penalty0 35549--35562, 2022.

\bibitem[Zhou et~al.(2018)Zhou, Xu, and Corso]{zhou2018towards}
Luowei Zhou, Chenliang Xu, and Jason Corso.
\newblock Towards automatic learning of procedures from web instructional videos.
\newblock In \emph{Proceedings of the AAAI Conference on Artificial Intelligence}, 2018.

\bibitem[Zohar et~al.(2024)Zohar, Wang, Dubois, Mehta, Xiao, Hansen-Estruch, Yu, Wang, Juefei-Xu, Zhang, et~al.]{zohar2024apollo}
Orr Zohar, Xiaohan Wang, Yann Dubois, Nikhil Mehta, Tong Xiao, Philippe Hansen-Estruch, Licheng Yu, Xiaofang Wang, Felix Juefei-Xu, Ning Zhang, et~al.
\newblock Apollo: An exploration of video understanding in large multimodal models.
\newblock \emph{arXiv preprint arXiv:2412.10360}, 2024.

\end{thebibliography}
}

\clearpage
\appendix

\section*{Appendix Overview}

The appendix presents more details and additional results not included in the main paper due to page limitations. The list of items included is:

\begin{compactitem}
\item Limitation in Section $\S$\ref{app:limitation}.
\item Ethic Statement results in Section $\S$\ref{app:ethic_statement}.
\item Extended Dataset Details in Section $\S$\ref{sec:data}.
\item Extended Details of the Dr.V-Agent in Section $\S$\ref{sec:Dr.V-Agent}.
\item  Extended Experiment Results and Analyses in Section $\S$\ref{sec:Extended Experiment Results and Analyses}.
\end{compactitem}

\section{Limitation}
\label{app:limitation}

While our design of \texttt{Dr.V-Agent} and \texttt{Dr.V-Bench} provides a fine-grained and interpretable framework for video hallucination diagnosis, we acknowledge several limitations in the current work.

\paratitle{Reliance on External Tool Performance.}
The effectiveness of \texttt{Dr.V-Agent} heavily relies on the performance of the external tools it invokes for tasks like spatial and temporal grounding. Although we employ SOTA tools, their inherent limitations can impact the accuracy of spatial-temporal annotations. Consequently, this can affect the precision of the final diagnosis and mitigation of hallucinations.

\paratitle{System Complexity and Computational Overhead.}
To achieve fine-grained reasoning, our multi-agent approach invokes up to eight external models, which introduces significant system complexity and computational overhead. Compared to end-to-end models, this multi-step, sequential process can lead to higher latency.

\paratitle{Annotation Cost and Scalability of the Benchmark.}
The construction of \texttt{Dr.V-Bench} depends on fine-grained manual spatial-temporal annotations. While this level of detail is critical for accurate hallucination diagnosis, the associated high annotation cost and time investment are key factors limiting the benchmark's scalability. This makes it challenging to rapidly expand the dataset to a larger scale or to more diverse video domains.

\paratitle{Indirect Assessment of Generative Capabilities.}
While \texttt{Dr.V-Bench} provides a robust benchmark for diagnosing hallucination, its evaluation of generative capabilities is adapted into a QA-based framework, specifically through ``caption generation QA". This design simplifies the targeted assessment of hallucinations but sacrifices a direct and comprehensive evaluation of a model’s free-form generative abilities, such as fluency, coherence, and creativity.

\section{Ethic Statement}
\label{app:ethic_statement}

The source videos of the \texttt{Dr.V-Bench} rely exclusively on publicly available video datasets. No extra private or sensitive content is used, ensuring compliance with ethical guidelines and respect for data privacy. 
The development of the spatial-temporal information is made possible through the dedicated efforts of a group of human annotators. 
All annotators are PhD students with specialized training in spatial-temporal annotation to ensure accuracy and reliability. 
In alignment with our goal of fostering an inclusive and collaborative research community, all data and resources of \texttt{Dr.V-Bench} will be openly available. This ensures equitable access for researchers and practitioners from diverse backgrounds, enabling broader contributions to advancements in video hallucination diagnosis and mitigation within LVMs.

\begin{table*}[!t]
\centering
\label{tab:comparison}
\resizebox{\linewidth}{!}{%
\begin{tabular}{@{}lcccccccccc@{}}
\toprule
\multirow{2}{*}{\textbf{Benchmark}} & \multirow{2}{*}{\textbf{Level}} & \multirow{2}{*}{\textbf{\#H-Types}} & \multirow{2}{*}{\textbf{\#Samples}} & \multirow{2}{*}{\textbf{\#Videos}} & \multirow{2}{*}{\textbf{Dur.(s)}} & \multirow{2}{*}{\textbf{\#Domains}} & \multirow{2}{*}{\textbf{Task}} &
\multirow{2}{*}{\textbf{S-T}} &
\multirow{2}{*}{\textbf{Interp.}}\\ 
 &  &  &  &  & & & & & \\ 
\midrule
SoraDetector \cite{chu2024sora}  & P, T  & 3  & 50  & 50 & [8, 20] & 6 & Open-Ended & \textcolor{red}{\ding{55}} & \textcolor{red}{\ding{55}} \\ 
VidHalluc \cite{li2024vidhalluc} & T &3  & 9,295 & 5,002 & [0, 220] & 3 & MCQ, Binary QA, Open-ended & \textcolor{red}{\ding{55}} & \textcolor{red}{\ding{55}} \\ 
VidHal \cite{choong2024vidhal}  & T &5 & 1,000 & 1,000 & [0, 85] & 9 & MCQ  & \textcolor{red}{\ding{55}} & \textcolor{red}{\ding{55}}\\ 
Eventhallusion \cite{zhang2024eventhallusion} & T &2 & 711 & 400 & [0, 30] & 7 & Binary QA, Open-Ended & \textcolor{red}{\ding{55}} & \textcolor{red}{\ding{55}}\\ 
VideoHallucer \cite{wang2024videohallucer} & P, T, C &5  & 1,800 & 948 & [7, 187] & 7 & Binary QA & \textcolor{red}{\ding{55}} & \textcolor{red}{\ding{55}}\\
ViBe \cite{rawte2024vibe}  & P, T & 5 &3,782  &3,782& [1, 2] & 5 & Open-Ended & \textcolor{red}{\ding{55}}  
 & \textcolor{red}{\ding{55}}\\ 

\textbf{Dr.V-Bench(Ours)} & P, T, C &14 & 10,000 & 4,974 & [0, 600] & 18 & MCQ, Binary QA, Open-Ended & \textcolor{green}{\ding{51}}  & \textcolor{green}{\ding{51}} \\
\bottomrule
\end{tabular}
}
\vspace{-2mm}
\caption{Comparison between our \textbf{\texttt{Dr.V-Bench}} and existing video hallucination benchmarks. We use the following abbreviations for conciseness: \textbf{Level} (P: Perceptive, T: Temporal, C: Cognitive), \textbf{\# H-Types} (Number of Hallucination Types), \textbf{S-T} (Spatial-Temporal Grounding), and \textbf{Interp.} (Interpretability).}
\end{table*}

\begin{table*}[ht]
\centering
\fontsize{8.5}{10}\selectfont
\setlength{\tabcolsep}{1.5mm}
\begin{tabular}{@{}llccc@{}}
\toprule
\textbf{Question Type} & \textbf{Hallucination Type} & \textbf{\# Samples} & \textbf{\# Videos} & \textbf{Data Source} \\ \midrule
\multirow{14}{*}{MCQ} & Object & 720 & 561 & STAR \\
 & Number & 300 & 296 & ActivityNet-QA, NExT-QA \\
 & Color & 300 & 295 & ActivityNet-QA \\
 & Location & 300 & 287 & NExT-QA \\
 & Static Relation & 480 & 480 & ActivityNet-QA \\
 & OCR & 300 & 180 & ViteVQA, RoadTextVQA \\
 & Action & 720 & 492 & STAR \\
 & Dynamic Attribute & 180 & 163 & TempCompass \\
 & Dynamic Relation & 900 & 634 & NExT-QA, STAR \\
 & Sequence & 120 & 117 & YouCook2 \\ 
 & Factual Prediction & 420 & 257 & STAR\\
 & Counterfactual Prediction & 300 & 300 & Causal-VidQA\\
 & Context-based Explanation & 600 & 382 & NExT-QA \\ 
 & Knowledge-based Explanation & 360 & 342 & Causal-VidQA, MMWorld, VCGBench-Diverse \\ \midrule
\multirow{14}{*}{Yes/No} & Object & 360 & 261 & STAR \\
 & Number & 150 & 150 & ActivityNet-QA, NExT-QA \\
 & Color & 150 & 123 & ActivityNet-QA, CLEVRER \\
 & Location & 150 & 150 & NExT-QA \\
 & Static Relation & 240 & 240 & ActivityNet-QA \\
 & OCR & 150 & 138 & ViteVQA \\
 & Action & 360 & 274 & STAR \\
 & Dynamic Attribute & 90 & 90 & TempCompass, Tomato \\
 & Dynamic Relation & 450 & 397 & NExT-QA, STAR, VCGBench-Diverse \\
 & Sequence & 60 & 59 & Video-MME, YouCook2 \\ 
  & Factual Prediction & 210 & 155 & STAR\\
 & Counterfactual Prediction & 150 & 150  & Causal-VidQA \\
 & Context-based Explanation & 300 & 255 & NExT-QA \\ 
 & Knowledge-based Explanation & 180 & 166 & Causal-VidQA, MMWorld, VCGBench-Diverse \\ \midrule
 \multirow{14}{*}{Caption} & Object & 120 & 120 & MSR-VTT, VATEX \\
 & Number & 50 & 50 & MSR-VTT, VATEX \\
 & Color & 50 & 50 & MSR-VTT, VATEX \\
 & Location & 50 & 50 & MSR-VTT, VATEX \\
 & Static Relation & 80 & 70 & MSR-VTT, VATEX \\
 & OCR & 50 & 49 & ViteVQA \\
 & Action & 120 & 120 & MSVD, MSR-VTT, VATEX \\
 & Dynamic Attribute & 30 & 16 & TempCompass \\
 & Dynamic Relation & 150 & 150 & MSR-VTT, VATEX \\
 & Sequence & 20 & 20 & YouCook2 \\  
  & Factual Prediction & 70 & 70 & MSR-VTT \\
 & Counterfactual Prediction & 50 & 50 & MSR-VTT \\
 & Context-based Explanation & 100 & 100 & MSR-VTT\\ 
 & Knowledge-based Explanation  & 60 & 60 & MSR-VTT\\ \bottomrule
\end{tabular}
\caption{Statistics of hallucination types across three categories of question-answering tasks, with a total of 10,000 samples and 4,974 videos in \texttt{Dr.V-Bench}.}
\label{tab:datasource}
\vspace{-4mm}
\end{table*}

\begin{table*}[!t]
\centering
\renewcommand{\arraystretch}{1.5}
\resizebox{\linewidth}{!}{
\begin{tabular}{m{2.5cm} m{3cm} m{2.5cm} m{13cm}}
\hline
\centering\textbf{Level} & \centering\textbf{Category} & \centering\textbf{Source} & \textbf{Example} \\ \hline
\multirow{7}{*}{\hspace{20pt}\textbf{Perception}} & \centering Object & \centering STAR & \textcolor{customcolor}{Which object was taken by the person?} (A) The clothes. (B) The shoe. (C) The broom. (D) The towel. \\ 
& \centering Number & \centering Activitynet-qa  NExT-QA & \textcolor{customcolor}{How many ducks in the video?} (A) one (B) three (C) two (D) five \\ 
& \centering Color & \centering Activitynet-qa  & \textcolor{customcolor}{What color is the car driving in the video?} (A) blue (B) red (C) yellow (D) green \\ 
& \centering Location & \centering NExT-QA   & \textcolor{customcolor}{Where is this video taken?} (A) market (B) forest (C) beach (D) cafe \\ 
& \centering Static Relation & \centering Activitynet-qa  & \textcolor{customcolor}{What is on the left of the river?} (A) bench (B) tree (C) house (D) fountain \\ 
& \centering OCR & \centering ViteVQA RoadTextVQA  & \textcolor{customcolor}{What is the price of the item?} (A) 124.99 (B) 134.99 (C) 149.99 (D) 139.89 \\ \hline

\multirow{6}{*}{\hspace{20pt}\textbf{Temporal}} & \centering Action & \centering STAR & \textcolor{customcolor}{What did the person do with the bag?} (A) Opened (B) Took (C) Put down (D) Threw \\ 
& \centering Dynamic Attribute & \centering TempCompass & \textcolor{customcolor}{In which direction are the women athletes running?} (A) from right to left (B) in circles (C) from left to right \\ 
& \centering Dynamic Relation & \centering STAR, NExT-QA & \textcolor{customcolor}{What did the lady with long hair do before she slid off?} (A) go back up (B) tie her hair (C) jump around (D) dance \\ 
& \centering Sequence & \centering YouCook2  & \textcolor{customcolor}{According to this video, in which order are the following items used for magic?} (a) Cards. (b) A stick. (c) A ring. (d) Paper money. (A) a, d, b, c (B) b, a, d, c (C) b, c, d, a (D) c, b, a, d \\ \hline

\multirow{6}{*}{\hspace{20pt}\textbf{Cognition}} & \centering Factual Prediction & \centering STAR & \textcolor{customcolor}{What will the person do next?} (A) Throw the box. (B) Take the bag. (C) Lie on the bed. (D) Sit on the floor. \\ 
& \centering Counterfactual Prediction & \centering Causal-VidQA & \textcolor{customcolor}{What will happen if the chair breaks suddenly?} (A) Maybe the wind will blow through the window. (B) The person will fall onto the ground. (C) The person can't eat two donuts; they can only eat one. (D) Students will be confused. \\ 
& \centering Context-based Explanation & \centering NExT-QA & \textcolor{customcolor}{Why did all the babies look at the blonde boy when he walked towards the man?} (A) babies are bored (B) man call him (C) curious (D) blonde boy is dancing \\ 
& \centering Knowledge-based Explanation & \centering Causal-VidQA MMWorld VCGBench-Diverse  & \textcolor{customcolor}{How does the left team win a point in the video?} (A) They make a free throw. (B) They score a layup. (C) They get an offensive rebound and a put-back. (D) They make a basketball 3-point shot. \\ \hline
\end{tabular}
}
\vspace{-2mm}
\caption{Multiple-choice QA examples of Dr.V-Bench's data sources.}
\label{qa_examples}
\end{table*}

\begin{table*}[!t]
    \centering
    \fontsize{7}{10}\selectfont
    \begin{tabular}{p{7cm} | p{9cm}}
        \toprule
        \textbf{Original QA} & \textbf{Generated QA} \\
        \midrule
        \multicolumn{2}{c}{\textit{\textbf{{Multiple-Choice QA}}}} \\
        \midrule
        \textbf{Hallucination Type:} Static Relation Hallucination & \textbf{Hallucination Type:} Static Relation Hallucination \\
        \textbf{Question:} Where is the lamp positioned in the scene? & \textbf{Question:} Where is the lamp positioned in the scene? \\
        \textbf{Choices:} A) Next to the couch. B) Behind the television. C) Above the sofa. D) The man is holding the lamp. & \textbf{Choices:} A) On the table. B) Under the sofa. C) Above the sofa. D) Adjacent to the window. \\
        \textbf{Answer:} C) Above the sofa. & \textbf{Answer:} C) Above the sofa. \\
        \midrule
        \multicolumn{2}{c}{\textit{\textbf{{Yes/No QA}}}} \\
        \midrule
        \textbf{Hallucination Type:} Object Hallucination & \textbf{Hallucination Type:} Object Hallucination \\
        \textbf{Question:} What object did the person put down? 
        & \textbf{Question:} Did the person put down the pillow? \\
        \textbf{Choices:} A) Hat. B) Skateboard. C) Sandwich. D) Pillow. & \textbf{Answer:} Yes. \\
        \textbf{Correct Answer:} D) Pillow. \\
        \midrule
        \multicolumn{2}{c}{\textit{\textbf{{Caption Generation QA}}}} \\
        \midrule
        \textbf{Hallucination Type:} Action Hallucination & \textbf{Hallucination Type:} Action Hallucination \\
        \textbf{Original Caption:} The video shows a person showing a red car's performance on a mountain road. & 
        \textbf{Choices:} \\
        & A: \{ ``objects": [``person", ``car"], ``relationship": ``drive", ``attributes": \{``car": ``red"\} \} \\
        & B: \{ ``objects": [``person", ``car"], ``relationship": ``show", ``attributes": \{``car": ``red"\} \} \\
        & C: \{ ``objects": [``person", ``car"], ``relationship": ``clean", ``attributes": \{``car": ``red"\} \} \\
        & D: \{ ``objects": [``person", ``car"], ``relationship": ``close", ``attributes": \{``car": ``red"\} \} \\
        \bottomrule
    \end{tabular}
    \vspace{-2mm}
    \caption{Examples of original and our generated QA.}
    \label{tab:transformedqa}
\end{table*}

\begin{figure*}[!t]
 \centering
 \includegraphics[width=1.0\linewidth]{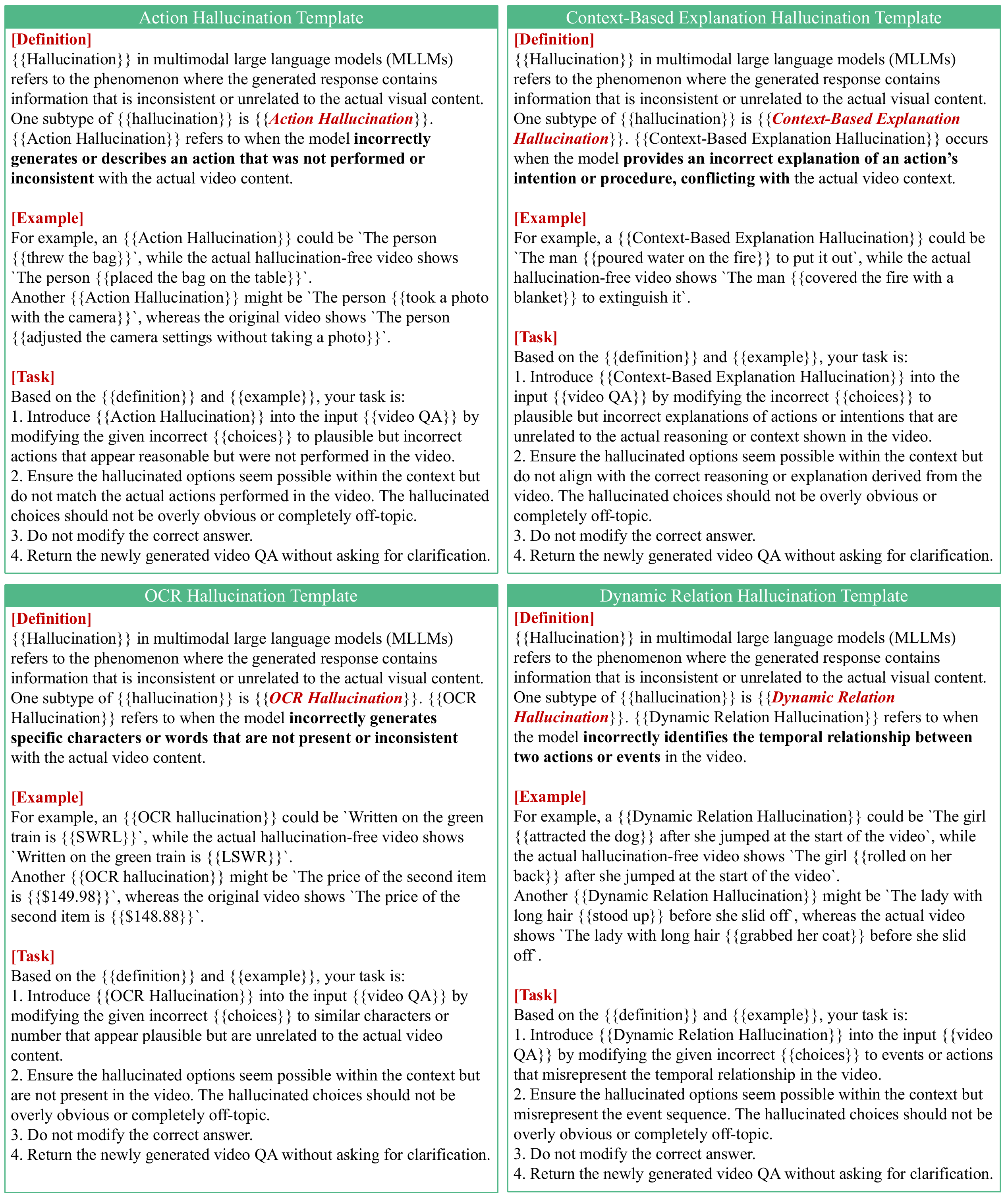}
\vspace{-2mm}
 \caption{Illustration of our designed prompts to generate multiple-choice QA.}
 \label{fig:prompt1}
\end{figure*}

\begin{figure*}[!t]
 \centering
 \includegraphics[width=0.90\linewidth]{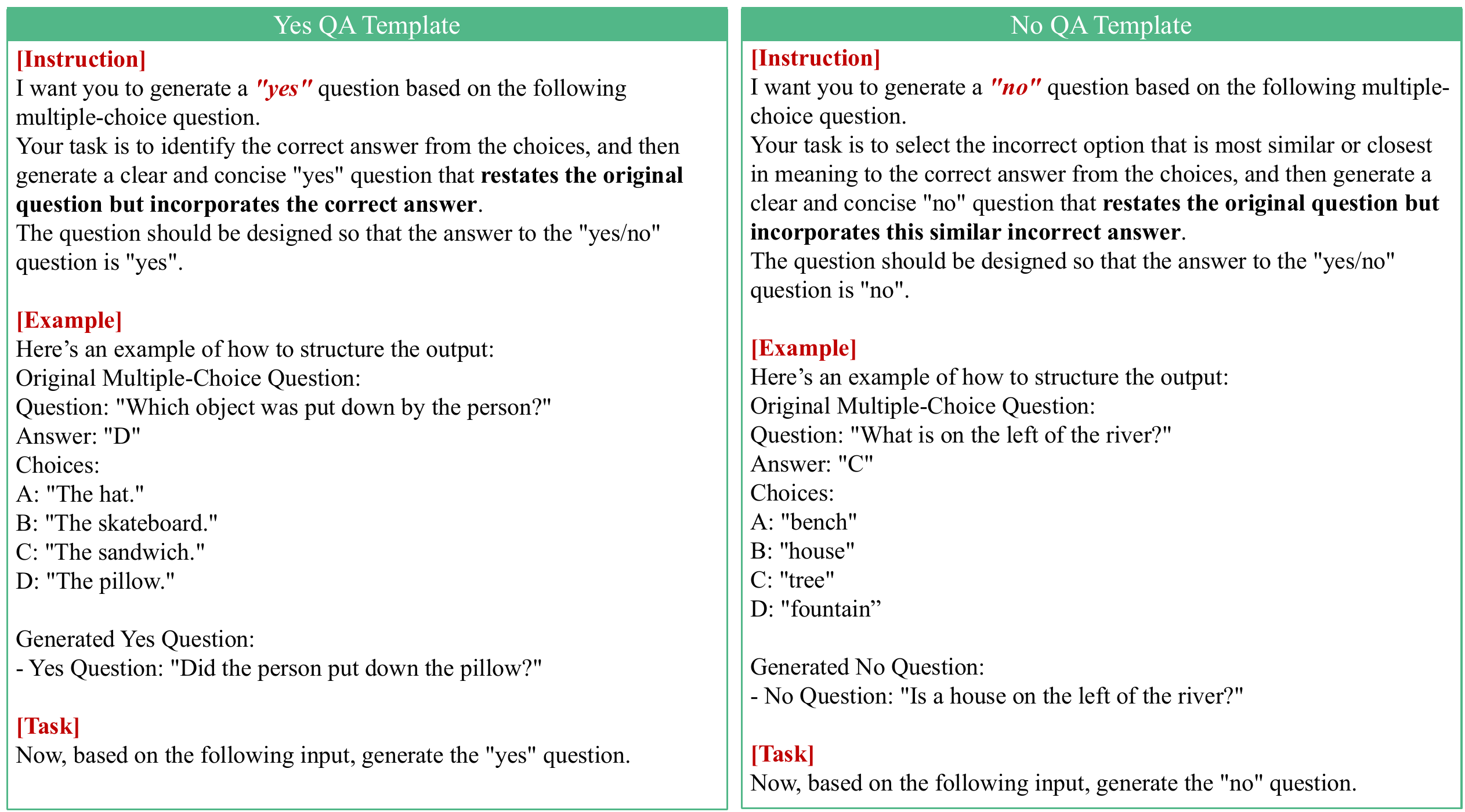}
 \vspace{-2mm}
 \caption{Illustration of our designed prompts to generate yes/no QA.}
 \label{fig:promptyesno}
\end{figure*}

\begin{figure*}[!t]
 \centering
\includegraphics[width=0.8\linewidth]{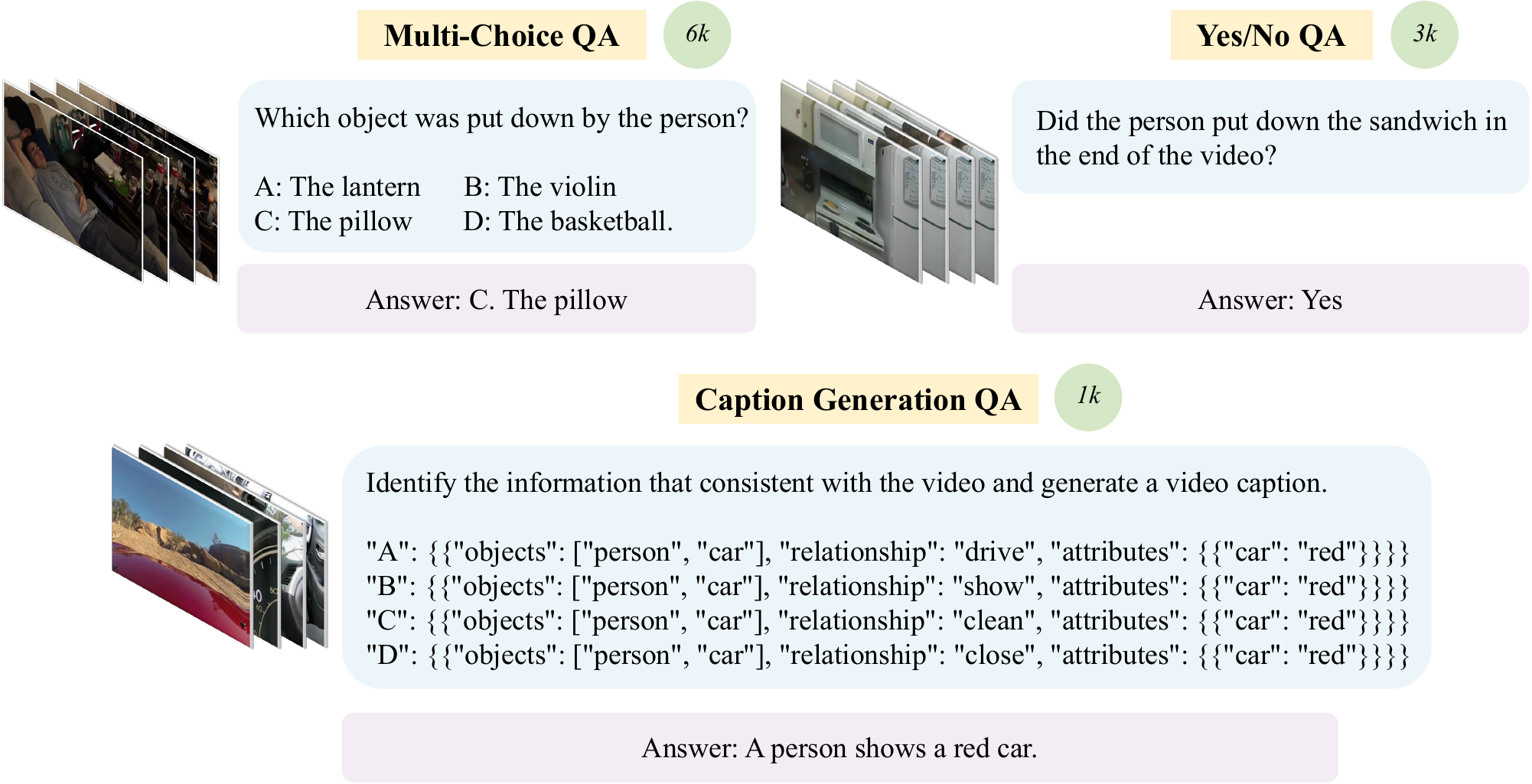}
\vspace{-2mm}
 \caption{Illustration of the three different types of QA tasks.}
 \label{task}
\end{figure*}

\section{Extended Dataset Details}
\label{sec:data}
The foundation of \texttt{Dr.V-Bench} is built upon a diverse collection of 15 well-established, public video datasets. These include ActivityNet-QA \cite{yu2019activitynet}, Causal-VidQA \cite{li2022representation}, NExT-QA \cite{xiao2021next}, YouCook2 \cite{zhou2018towards}, CLEVRER \cite{yi2019clevrer}, ViteVQA \cite{zhao2022towards}, RoadTextVQA \cite{tom2023reading}, TempCompass \cite{liu2024tempcompass}, Tomato \cite{shangguan2024tomato}, MMWorld \cite{he2024mmworld}, Video-MME \cite{fu2024video}, VCGBench-Diverse \cite{maaz2024videogpt}, MSVD \cite{chen2011collecting}, MSR-VTT \cite{xu2016msr}, and VATEX \cite{wang2019vatex}. As summarized in Table \ref{tab:datasource}., these sources cover a wide array of video types and tasks, providing a robust basis for evaluation.

\subsection{Taxonomy of Video Hallucination}

\begin{compactitem}
    \item \textbf{Perceptive Hallucination}. This level of hallucination involves the model's failure to correctly recognize basic static content in the video, including object recognition, identification of static attributes such as number, color, and position, understanding static spatial relations between objects, and extracting textual information through processes like OCR. This is the lowest level of hallucination and is the stage at which many LVMs commonly exhibit errors.

    \item \textbf{Temporal Hallucination.} Temporal hallucinations involve deficiencies in a model's capability to process and interpret dynamic temporal information within videos. LVMs typically struggle with recognizing actions and events, identifying dynamic attributes such as the direction of movement, speed, and sequences of actions, understanding the dynamic relations between events, and recognizing the action sequence of the video.  

    \item \textbf{Cognitive Hallucination.} Cognitive hallucinations encompass failures in the semantic understanding and cognitive processing of video content. LVMs exhibiting cognitive hallucination often fail in several key aspects. First, they may struggle with factual prediction, which requires predicting future actions beyond the content of the current video clips. Second, they may exhibit errors in counterfactual prediction, which involves imagining what would happen under hypothetical or different conditions. Third, they often face challenges with context-based explanation, which requires understanding the video context to explain the intentions behind actions and the processes that lead to specific outcomes. Finally, they may struggle with knowledge-based explanation, which demands commonsense knowledge or domain-specific expertise to interpret and answer questions.  
\end{compactitem}

\subsection{Comparison with Existing Benchmarks}
Our \texttt{Dr.V-Bench} represents significant advancements over existing video hallucination benchmarks. 

\paratitle{Taxonomy depth.} \texttt{Dr.V-Bench}
has the most detailed video hallucination taxonomy to date, organizing hallucinations into three hierarchical levels with 14 distinct types.

\paratitle{Task diversity.} \texttt{Dr.V-Bench} incorporates diverse task formats, i.e., yes/no QA, multiple-choice QA, and captioning task that covers hallucinations across all possible video tasks.

\paratitle{Spatial-temporal grounding.} Our benchmark is the only one to provide fine-grained spatial-temporal annotations with bbox coordinates in all related frames, from start to end timestamps, that capture crucial moments of object interaction or movement,  for understanding context and diagnosing hallucinations accurately.

\paratitle{Temporal Coverage and Duration Diversity.}
\texttt{Dr.V-Bench} offers unparalleled diversity in video duration, with a range spanning from short clips to long videos of up to 600 seconds. This vast range is critical for robust model evaluation and stands in stark contrast to other benchmarks that are often limited to much shorter videos. 
By including both short and extremely long videos, our benchmark rigorously tests a model's ability to handle varying temporal scales, from capturing fleeting actions to understanding long-range narrative dependencies.

\paratitle{Domain and Scenario Richness.}
\texttt{Dr.V-Bench} provides the most extensive domain coverage to date, encompassing 18 distinct domains. This number significantly surpasses that of all other existing benchmarks, which typically cover fewer than 10 domains. 
This richness ensures that models are evaluated across a wide array of real-world contexts and scenarios, preventing overfitting to specific visual styles or topics and providing a true test of their generalization capabilities.

\paratitle{Dataset scale.} The scale of \texttt{Dr.V-Bench} significantly surpasses that of other datasets. Its extensive coverage in terms of instances, unique videos, and diverse domains, combined with a high proportion of complex scenarios, provides a more challenging and comprehensive evaluation platform. This ensures that models are tested on their ability to generalize and perform higher-level reasoning, pushing the boundaries beyond existing benchmarks.

\subsection{Hallucination-centric QA Evaluation}
The yes/no and multiple-choice tasks are considered discriminative hallucination evaluation, focusing on whether a model can correctly identify the ground truth among plausible but incorrect options. In contrast, the caption generation QA introduces a generative hallucination evaluation.
For this task, we provide the model with structured information derived from the question and instruct it to select the correct details to generate an appropriate video caption. This constrained format facilitates a more focused evaluation of captioning capabilities and simplifies automated assessment. Concrete examples of these transformations are shown in Table \ref{tab:transformedqa}.
The evaluation protocol is then tailored to each task type:
\begin{compactitem}
    \item \textbf{For Yes/No and Multiple-Choice QA}, we first check if the model's response explicitly includes a candidate option (e.g., 'Yes', 'No', 'A', 'B', 'C', or 'D') that matches the ground truth. If no direct match is found, we then utilize GPT-4o's language understanding capabilities to determine if the response semantically aligns with the correct answer.

    \item \textbf{For Caption Generation QA}, rule-based evaluation is ineffective due to the free-form nature of captions. Therefore, we rely exclusively on GPT-4o to evaluate the accuracy of the generated captions, ensuring the output faithfully reflects the structured information provided and the video's content.
\end{compactitem}

\subsection{Fine-grained Spatial-Temporal Grounding}
\label{sec:label}

\begin{compactenum}
    \item \textbf{Extraction of Target Objects.} In this step, the key target objects required for the QA are identified. These objects are crucial for the subsequent annotation process. The selection of target objects is based on their relevance to the question and their presence in the video content.

    \item \textbf{Annotation of Start and End Frames.} Each target object is annotated with its start and end frames. The start frame is defined as the frame where at least 30\% of the object's contours first appear, and the end frame is defined as the frame where at least 30\% of the object's contours disappear. This ensures accurate capture of the object’s temporal boundaries.

    \item \textbf{Annotation of Key Frames.} Within the interval defined by the start and end frames, key frames for each target object are annotated. Key frames are those that effectively aid in the recognition and answering of the QA. These frames typically depict significant changes in the object's state, position, or interaction with other elements in the scene.

    \item \textbf{Annotation of Bounding Boxes.} After annotating the start, end, and key frames, bounding boxes (bbox) are annotated for each frame. This involves marking the spatial location of each target object in every frame. The bounding box (bbox) coordinates are recorded to provide precise spatial information for the target objects.
\end{compactenum}

\subsection{Quality Control}
To ensure the highest quality and consistency, we implement a strict quality control protocol. All annotators are PhD students from computer science fields who receive standardized training and use a unified annotation platform. Each video is independently annotated by two annotators. We measure inter-annotator agreement (IAA) quantitatively, achieving a high score of 0.85. This is calculated using the average Intersection over Union (IoU) for bounding boxes and temporal overlap for keyframes. Any samples falling below this agreement threshold or showing significant discrepancies are either resolved by a senior annotator or discarded from the final dataset.

\begin{figure*}[!t]
 \centering
 \includegraphics[width=0.80\linewidth]{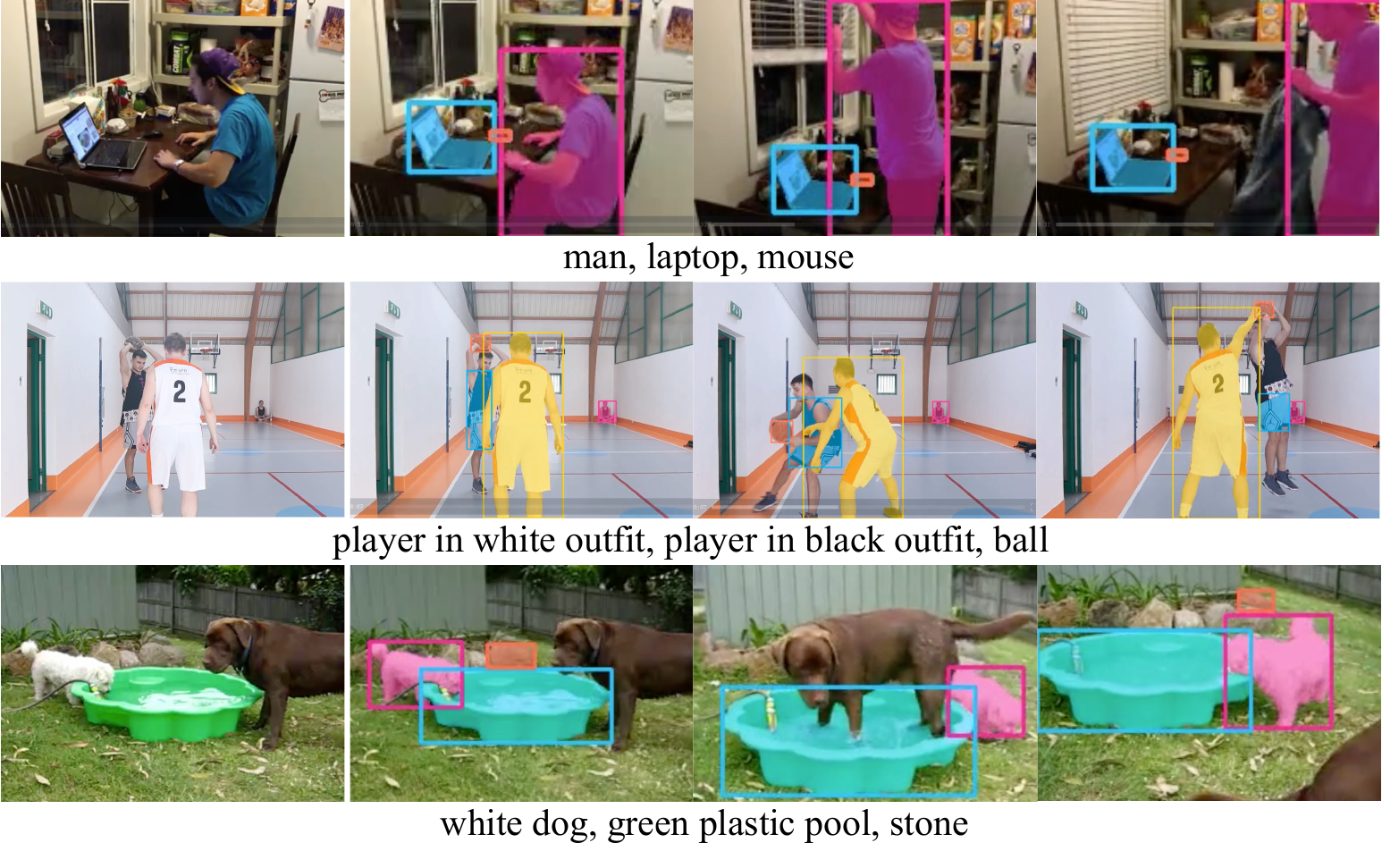}
   \vspace{-3mm}
 \caption{Video object tracking results using Grounded SAM2.}
 \label{fig:sam2}
\end{figure*}

\begin{figure*}[!t] 
 \centering
 \includegraphics[width=0.80\linewidth]{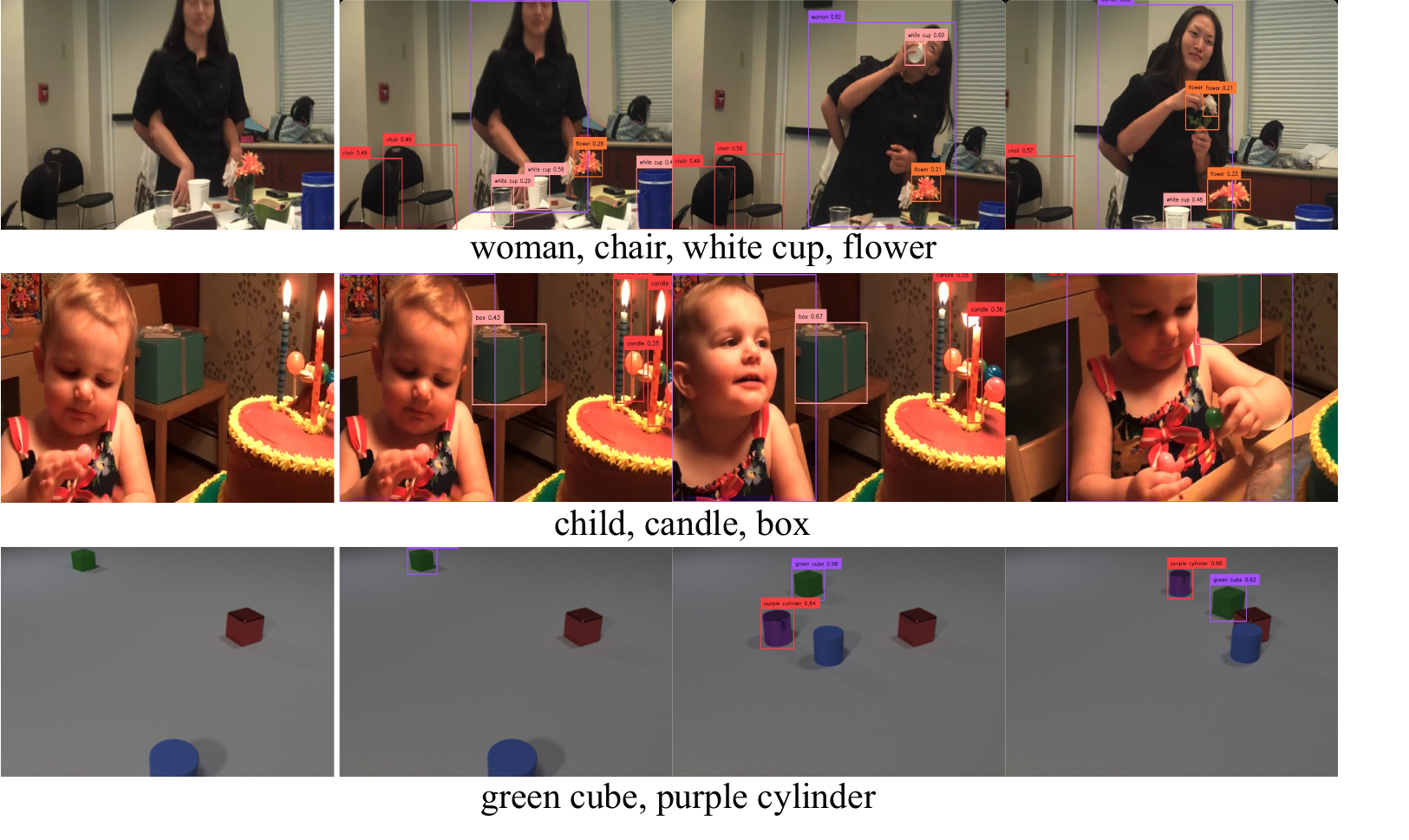}
  \vspace{-3mm}
 \caption{Video object tracking results using YOLO-World.}
 \label{fig:yoloworld}
\end{figure*}

\begin{figure*}[!t] 
 \centering
 \includegraphics[width=0.85\linewidth]{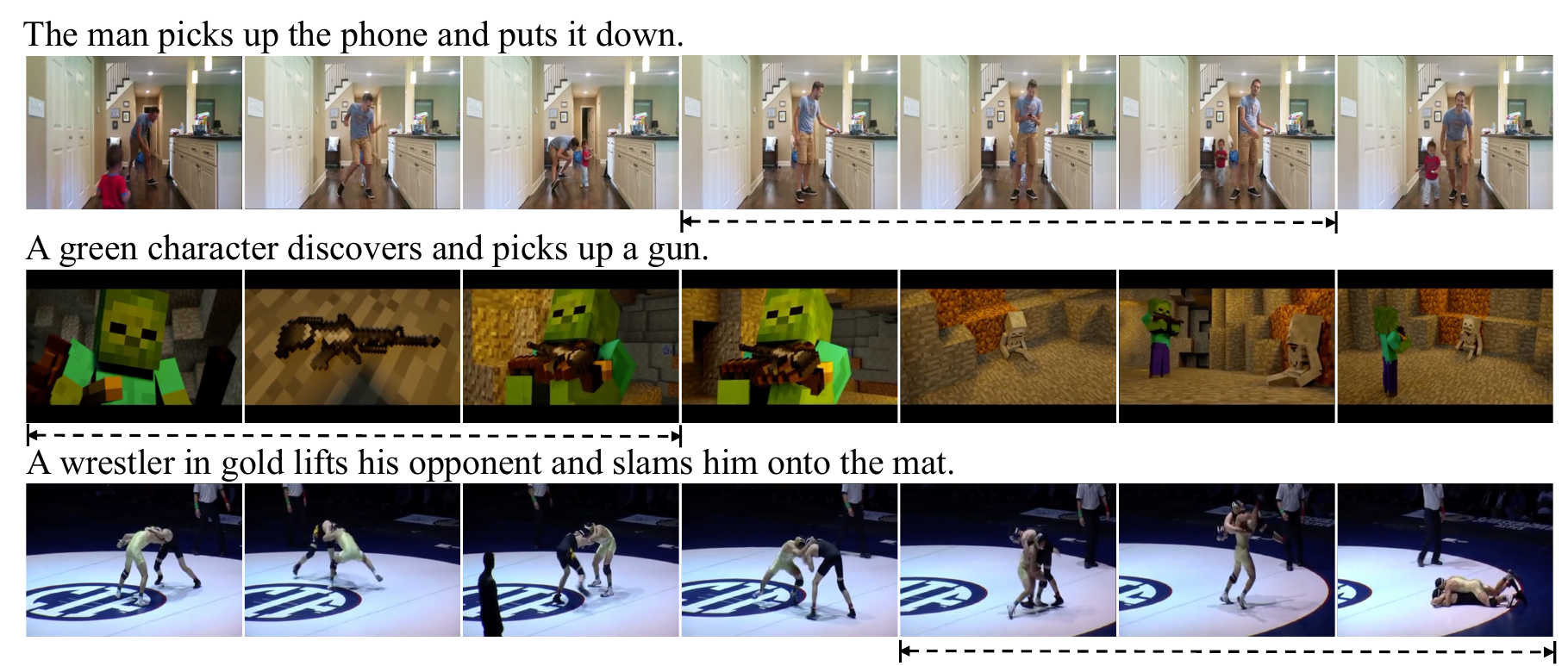}
 \vspace{-3mm}
 \caption{Video temporal grounding results using CG-STVG.}
 \label{fig:cg}
\end{figure*}

\begin{figure*}[!t]
 \centering
 \includegraphics[width=0.85\linewidth]{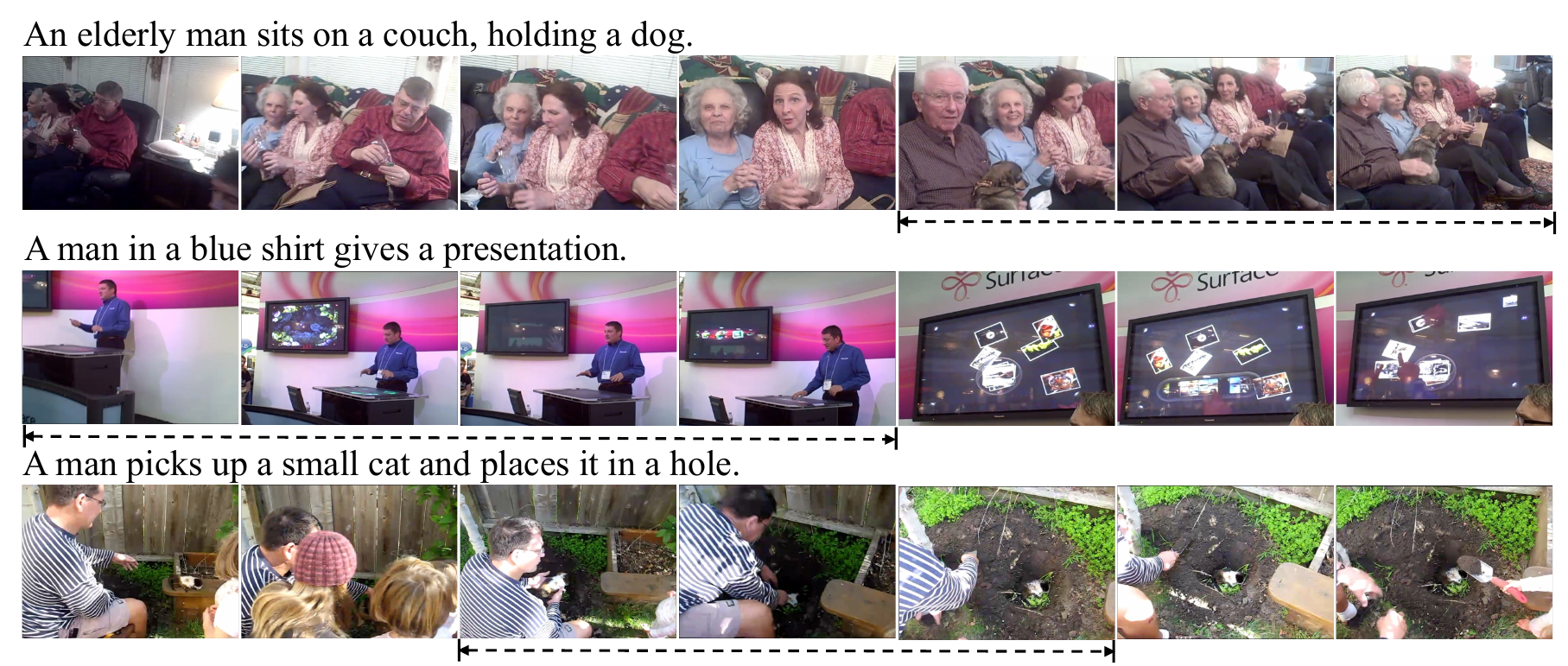}
  \vspace{-3mm}
 \caption{Video temporal grounding results using Grounded-VideoLLM.}
 \label{fig:ground}
\end{figure*}

\section{Extended Details of the \DrVLogo \texttt{Dr.V-Agent}}
\label{sec:Dr.V-Agent}

\subsection{Open-Vocabulary Video Object Tracking}
\paratitle{Grounded SAM 2} \cite{ren2024grounded} builds on the robust segmentation and tracking capabilities of SAM 2 by integrating it with Grounding DINO \cite{liu2023grounding}. This combination enables open-set object segmentation and tracking within video content. By processing input prompts, the model can precisely track specified objects and provide their normalized location coordinates. This functionality allows our system to achieve accurate and flexible video object tracking across diverse scenarios (see Figure~\ref{fig:sam2} for results).

\paratitle{YOLO-World} \cite{Cheng2024YOLOWorld} is a next-generation YOLO detector designed for open-vocabulary object detection and grounding. Leveraging pre-training on large-scale datasets, it supports user-defined vocabulary prompts to detect and ground objects efficiently. This capability ensures robust object detection and localization, making it a vital tool for video object tracking in \texttt{Dr.V-Agent} (see Figure~\ref{fig:yoloworld} for results).

\subsection{Video Temporal Grounding}
\paratitle{Context-Guided Spatial-Temporal Video Grounding (CG-STVG)} \cite{gu2024context} is designed for precise video temporal grounding through a two-stage architecture: a multimodal encoder for feature extraction and a context-guided decoder for refined grounding. The model has achieved great performance on many challenging benchmarks, making it a highly reliable choice for temporal grounding tasks in our system (see Figure~\ref{fig:cg} for results).

\paratitle{Grounded-VideoLLM} \cite{wang2024grounded} is a specialized VLM optimized for fine-grained temporal grounding, demonstrating strong performance in temporal sentence grounding and grounded VideoQA tasks. The model incorporates a temporal stream that encodes inter-frame relationships using discrete temporal tokens enriched with time-specific knowledge, making it a versatile tool for general video understanding (see Figure~\ref{fig:ground} for results).

\section{Extended Experiment Results and Analyses}
\label{sec:Extended Experiment Results and Analyses}

\subsection{In-depth Analysis of Hallucination Causes}
The performance data from Table~\ref{tab:all types} allows us to diagnose the likely reasons for model failures at each level of understanding. We identify consistent patterns of failure that hold true across nearly all evaluated LVMs.

\begin{table*}[!t]
    \centering
    \setlength{\tabcolsep}{1.6mm}
    \renewcommand{\arraystretch}{1.2}
    \resizebox{\linewidth}{!}{
        \begin{tabular}{lccccccccccccccc}
            \toprule
            \textbf{Model} & \textbf{Obj.} & \textbf{Num.} & \textbf{Col.} & \textbf{Loc.} & \textbf{SRel.} & \textbf{OCR} & \textbf{Act.} & \textbf{Atr.} & \textbf{DRel.} & \textbf{Seq.} & \textbf{Fct.} & \textbf{CnFct.} & \textbf{Cxt.} & \textbf{Knw.} & \textbf{Avg} \\ 
            \midrule
            \rowcolor[gray]{0.9} \multicolumn{16}{c}{\textit{\textbf{Multiple-Choice QA}}} \\
            \midrule
            VideoChat2 & 50.00 & 32.89 & 39.13 & 72.67 & 29.89 & \underline{29.00} & \underline{26.03} & \underline{25.00} & 27.84 & \underline{25.50} & 32.27 & \underline{28.36} & 33.67 & \underline{30.00} & 34.45 \\
            Video-ChatGPT & \underline{35.25} & \underline{30.87} & \underline{28.09} & \underline{67.00} & \underline{25.37} & 37.13 & 26.17 & 28.89 & \underline{25.67} & 26.23 & \underline{25.43} & 28.00 & \underline{32.00} & 34.64 & \underline{32.20} \\
            Video-LLaVA & 50.97 & 40.60 & 48.49 & 81.33 & 57.86 & 37.67 & 30.42 & 34.44 & 32.33 & 45.00 & 49.76 & 37.00 & 63.67 & 48.89 & 47.03 \\
            LLaMA-VID & 49.72 & 42.95 & 40.80 & 82.00 & 56.81 & 50.67 & 32.78 & 33.89 & 31.22 & 42.72 & 43.57 & 55.33 & 63.17 & 52.66 & 48.45 \\
            LLaVA-NeXT-Video-DPO & 49.58 & 68.46 & 53.51 & 85.00 & 62.47 & 44.33 & 40.28 & 38.89 & 40.56 & 29.17 & 45.24 & 61.67 & 73.00 & 55.00 & 53.37 \\
            PLLaVA & 67.92 & 72.82 & 67.56 & 85.50 & 77.36 & 54.67 & 36.94 & 46.67 & 50.78 & 40.00 & 65.24 & 68.67 & 74.33 & 61.94 & 62.17 \\
            InternVL2 & 76.67 & 71.48 & 73.58 & 92.33 & 79.04 & 64.67 & 59.86 & 45.56 & 67.11 & 55.83 & 69.52 & 64.67 & 80.00 & 64.80 & 68.94 \\
            Qwen2-VL & 79.00 & 75.51 & 72.58 & 91.00 & 79.08 & 75.33 & 66.03 & 52.22 & 73.67 & 70.50 & 70.86 & 73.33 & 85.83 & 71.74 & 74.05 \\
            GPT-4o & 83.64 & 80.53 & 78.42 & 93.50 & 83.25 & 80.32 & 74.42 & 58.12 & 78.64 & 75.53 & 75.32 & 78.05 & 91.43 & 76.12 & 79.09 \\
            Gemini-1.5-Pro & \textbf{85.31} & \textbf{83.42} & \textbf{81.32} & \textbf{94.09} & \textbf{85.56} & \textbf{82.52} & \textbf{77.63} & \textbf{61.13} & \textbf{81.52} & \textbf{77.13} & \textbf{78.14} & \textbf{81.04} & \textbf{93.53} & \textbf{79.42} & \textbf{81.55} \\
            \midrule
            \rowcolor[gray]{0.9} \multicolumn{16}{c} {\textit{\textbf{Yes/No QA}}} \\
            \midrule
            VideoChat2 & \underline{27.22} & \underline{22.67} & \underline{36.24} & \underline{50.67} & \underline{49.07} & \underline{31.33} & \underline{46.94} & \underline{35.56} & \underline{60.22} & 33.33 & 44.29 & 46.67 & \underline{40.33} & \underline{30.00} & \underline{39.61} \\
            Video-ChatGPT & 48.33 & 49.33 & 48.32 & 52.67 & 49.58 & 54.67 & 51.67 & 48.88 & 65.78 & 40.00 & 49.52 & 54.00 & 51.67 & 48.33 & 50.91 \\
            Video-LLaVA & 58.33 & 65.33 & 53.02 & 81.33 & 56.36 & 51.33 & 50.28 & 51.11 & 69.78 & 46.67 & 47.14 & \underline{42.67} & 57.00 & 51.11 & 55.82 \\
            LLaMA-VID & 59.44 & 54.00 & 50.34 & 74.67 & 57.63 & 55.33 & 48.89 & 52.22 & 67.56 & \underline{28.57} & 50.95 & 45.33 & 55.67 & 54.80 & 53.96 \\
            LLaVA-NeXT-Video-DPO & 61.67 & 75.33 & 64.43 & 81.33 & 63.56 & 56.67 & 60.00 & 53.33 & 65.11 & 55.00 & \underline{42.38} & 60.67 & 67.33 & 60.00 & 61.92 \\
            PLLaVA & 71.11 & 78.00 & 64.43 & 80.67 & 79.66 & 62.67 & 52.50 & 52.22 & 67.33 & 53.33 & 49.52 & 64.00 & 67.67 & 61.67 & 64.63 \\
            InternVL2 & 69.72 & 78.00 & 72.48 & 89.33 & 71.61 & 64.00 & 61.67 & 60.00 & 62.22 & 46.67 & 51.90 & 54.67 & 65.33 & 62.78 & 65.03 \\
            Qwen2-VL & 72.78 & 80.67 & 73.83 & 90.67 & 81.36 & 72.00 & 65.28 & 62.22 & 66.67 & 53.33 & 69.52 & 61.33 & 74.33 & 72.78 & 71.20 \\
            GPT-4o & 77.51 & 84.61 & 78.08 & 93.05 & 84.06 & 76.80 & 70.20 & 66.08 & 71.30 & 58.06 & 74.06 & 67.19 & 77.90 & 76.08 & 75.36 \\
            Gemini-1.5-Pro & \textbf{80.35} & \textbf{86.18} & \textbf{81.31} & \textbf{94.60} & \textbf{86.08} & \textbf{78.08} & \textbf{73.50} & \textbf{68.27} & \textbf{73.25} & \textbf{60.20} & \textbf{77.29} & \textbf{69.71} & \textbf{80.20} & \textbf{79.18} & \textbf{77.73} \\
            \midrule
            \rowcolor[gray]{0.9} \multicolumn{16}{c}{\textit{\textbf{Caption Generation QA}}} \\
            \midrule
            VideoChat2 & 37.14 & 36.00 & 38.04 & 54.14 & 35.40 & 37.00 & 43.56 & \underline{29.41} & \underline{29.96} & 32.20 & 40.30 & \underline{29.41} & 40.45 & 38.54 & 37.25 \\
            Video-ChatGPT & \underline{34.00} & \underline{29.81} & \underline{38.01} & \underline{46.17} & \underline{29.00} & \underline{30.00} & \underline{38.88} & \underline{29.41} & 30.51 & \underline{30.45} & \underline{35.00} & 33.56 & \underline{35.75} & \underline{37.64} & \underline{34.16} \\
            Video-LLaVA & 54.90 & 50.21 & 55.80 & 77.34 & 50.26 & 48.54 & 50.41 & 39.34 & 47.56 & 40.15 & 44.90 & 43.04 & 58.56 & 53.64 & 51.05 \\
            LLaMA-VID & 47.51 & 45.54 & 50.54 & 71.21 & 45.61 & 43.96 & 47.42 & 48.43 & 40.52 & 37.88 & 45.65 & 48.60 & 64.26 & 58.25 & 49.67 \\
            LLaVA-NeXT-Video-DPO & 65.15 & 63.93 & 65.82 & 82.50 & 68.35 & 56.53 & 60.21 & 54.54 & 56.00 & 50.00 & 59.05 & 55.62 & 66.45 & 64.56 & 62.05 \\
            PLLaVA & 59.14 & 59.35 & 59.34 & 79.56 & 63.31 & 59.35 & 56.90 & 53.33 & 52.63 & 45.05 & 57.12 & 52.53 & 66.34 & 60.64 & 58.90 \\
            InternVL2 & 69.61 & 66.64 & 69.30 & 87.30 & 72.34 & 62.00 & 62.45 & 56.24 & 59.10 & 53.32 & 61.88 & 60.16 & 69.47 & 66.64 & 65.46 \\
            Qwen2-VL & 73.35 & 71.50 & 71.44 & 85.32 & 74.53 & 67.54 & 65.21 & 58.12 & 63.44 & 57.25 & 66.34 & 64.32 & 74.15 & 70.32 & 68.77 \\
            GPT-4o & 78.08 & 74.60 & 75.57 & 90.09 & 76.38 & 72.77 & 68.70 & 61.36 & 66.28 & 60.20 & 69.17 & 67.55 & 77.97 & 73.50 & 72.30 \\
            Gemini-1.5-Pro & \textbf{80.02} & \textbf{76.48} & \textbf{77.19} & \textbf{92.02} & \textbf{78.08} & \textbf{74.05} & \textbf{71.30} & \textbf{63.50} & \textbf{68.70} & \textbf{62.04} & \textbf{71.30} & \textbf{69.25} & \textbf{80.60} & \textbf{76.04} & \textbf{74.33} \\
            \bottomrule
        \end{tabular}
    }
    \vspace{-2mm}
    \caption{Accuracy of LVMs across different hallucination types in three tasks.
    Columns represent the hallucination types: Object (Obj.), Color (Col.), Number (Num.), Location (Loc.), Static Relation (SRel.), OCR; Action (Act.), Dynamic Attribute (Atr.), Dynamic relation (DRel.), Sequence (Seq.); Factual Prediction (Fct.), Counterfactual Prediction (CnFct.), Context-based Explanation (Cxt.), Knowledge-based Explanation (Knk.).  The highest score is in \textbf{bold} and the lowest is \underline{underlined}.}
    \label{tab:all types}
\end{table*}

\begin{compactitem}
    \item \textbf{Perceptive Level: Capable in General, but Lacking Fine-Grained Fidelity.} 
    Although this is the highest-scoring category, a clear pattern emerges: models excel at coarse-grained tasks but falter on fine-grained details. This trend is universal across all tested models; for instance, performance on ``Location'' (e.g., Qwen2-VL: 91.00\%) and ``Object'' (e.g., InternVL2: 76.67\%) consistently surpasses that on detail-oriented tasks like ``Number'' (Qwen2-VL: 75.51\%) and ``Color'' (InternVL2: 73.58\%) in the Multiple-Choice QA task. This suggests that while core visual feature extraction is robust enough for general scene understanding, it lacks the high fidelity required for precise counting or distinguishing subtle attributes. Furthermore, ``OCR'' remains a distinct challenge, with average scores often lagging behind other perception tasks, indicating persistent difficulties in processing embedded text within visual scenes.

    \item \textbf{Temporal Level: A Fundamental Weakness in Modeling Dynamics.}
    This level marks a significant performance drop for all models, pointing to a fundamental weakness in modeling temporal dynamics. Across the board, the most challenging sub-tasks are consistently ``Dynamic Attribute'' and ``Sequence''. As a stark example from Table~\ref{tab:types}, the top-performing Gemini-1.5-Pro sees its accuracy drop to just 61.13\% on ``Dynamic Attribute'' (Atr.) in the Multiple-Choice QA---one of its lowest scores across all categories. This highlights a fundamental inability to track object state changes over time. Similarly, the universally low scores on ``Sequence'' reveal that current architectures, often adapted from static image models, are insufficient for capturing the complex, long-range dependencies and ordering of events inherent in video.

    \item \textbf{Cognitive Level: Failure in Abstract and Causal Reasoning.}
    This level represents the most significant challenge, consistently yielding the lowest average scores. Such failures are twofold. First, they are often a cascade effect from temporal errors, as seen when Video-ChatGPT's accuracy plummets from a respectable 67.00\% on the perceptive task ``Location'' to a mere 28.00\% on ``Counterfactual Prediction'' (CnFct.). Second, and more critically, models fail at abstract reasoning itself. A deeper look inside the cognitive tasks reveals a crucial distinction: models perform significantly worse on tasks requiring true reasoning—``Counterfactual Prediction'' and ``Knowledge-based Explanation''—than on those that can be partially solved with contextual cues (``Context-based Explanation''). For example, even Gemini-1.5-Pro scores a near-perfect 93.53\% on ``Context-based Explanation'' but drops to 81.04\% on ``Counterfactual Prediction'' and 79.42\% on ``Knowledge-based Explanation''. This demonstrates that the primary bottleneck lies not only in understanding the video’s context, but also in performing robust causal reasoning and integrating external world knowledge.
\end{compactitem}

\subsection{Evaluation of Dr.V-Agent's Spatial-Temporal Grounding}
\label{Evaluation of Spatial-Temporal Grounding}
We randomly sample 2,000 instances from Dr.V-Bench to evaluate the spatial and temporal grounding abilities of our chosen models. Following~\cite{gu2024context}, we adopt \text{m\_tIoU}, \text{m\_vIoU}, and \text{vIoU}@R as our standard evaluation metrics. Specifically, \text{m\_tIoU} measures temporal localization accuracy, computed as the mean temporal Intersection over Union (\text{tIoU}) across all test sequences, where \(\text{tIoU} = \frac{|\mathcal{P}_i|}{|\mathcal{P}_u|}\), with \( \mathcal{P}_i \) and \( \mathcal{P}_u \) denoting the temporal intersection and union of the predicted and ground-truth temporal segments. Similarly, \text{m\_vIoU} evaluates spatial localization by averaging the visual IoU (\text{vIoU}) over all test videos, where \(\text{vIoU} = \frac{1}{|\mathcal{P}_u|} \sum_{t \in \mathcal{P}_i} \text{IoU}(b^*_t, b_t)\), with \( b^*_t \) and \( b_t \) representing the ground-truth and predicted bounding boxes at frame \( t \). Lastly, \text{vIoU}@R represents the proportion of test samples where \text{vIoU} exceeds a predefined threshold \( R \). As shown in Table~\ref{context_tdb}, Grounded SAM 2 and YOLO-World excel in visual segmentation and object tracking, confirming their strong perceptive-level precision, while CG-STVG and Grounded-VideoLLM demonstrate robust temporal localization, which validates their ability to accurately detect and align events over time. These results highlight the effectiveness of the perceptive and temporal-level tools integrated into Dr.V-Agent, providing a solid foundation for mitigating video hallucinations in LVMs.

\begin{table}[ht]
\centering
\resizebox{\linewidth}{!}{
\begin{tabular}{lcccc}
\toprule
\textbf{Model} & \textbf{m\_tIoU} & \textbf{m\_vIoU} & \textbf{vIoU@0.3} & \textbf{vIoU@0.5} \\ 
\midrule
Grounded SAM 2      & 50.06 & 42.27 & 62.60 & 39.25 \\  
YOLO-World          & --    & 43.83 & 61.42 & 37.69 \\
CG-STVG             & 56.11 & 38.75 & 59.05 & 36.28 \\ 
Grounded-VideoLLM   & 53.90 & --    & --    & --    \\ 
\bottomrule
\end{tabular}
}
\vspace{-2mm}
\caption{Spatial-temporal grounding performance of each model used in \textbf{\texttt{Dr.V-Agent}}.}
\label{context_tdb}
\end{table}

\subsection{Impact of Question Format on Task Difficulty}
\begin{figure*}[!t]
 \centering
 \includegraphics[width=0.95\linewidth]{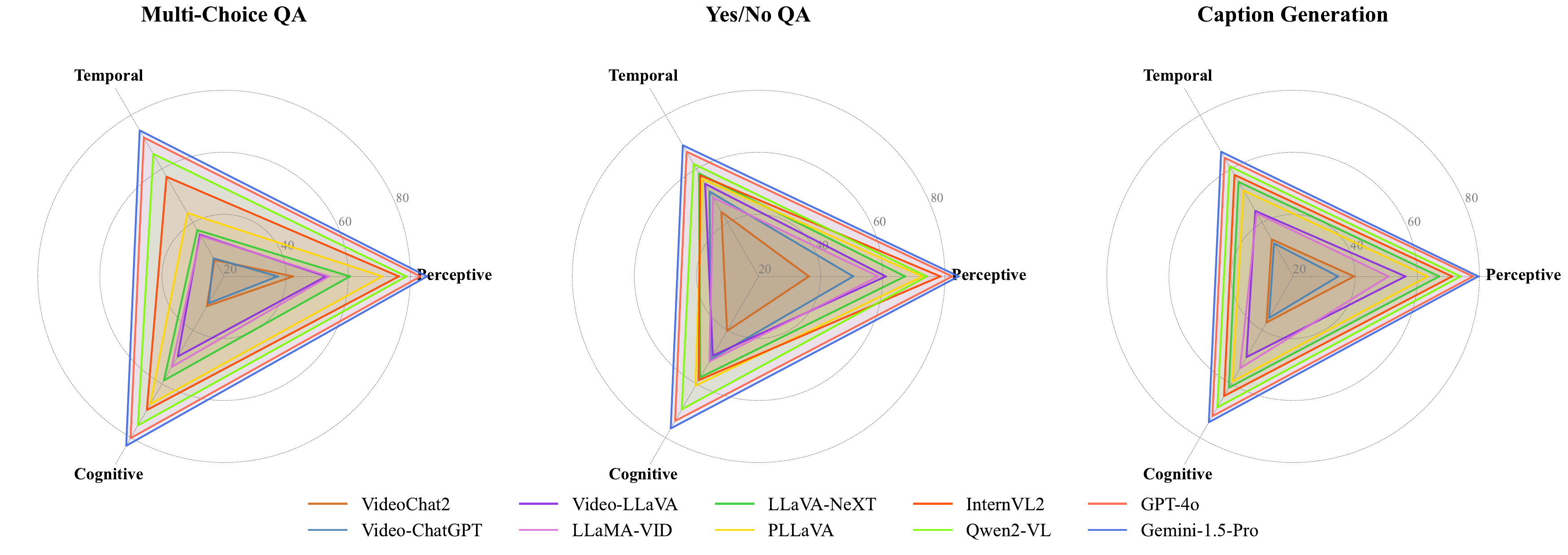}
 \vspace{-2mm}
 \caption{Accuracy of LVMs across different hallucination levels in three tasks.}
 \label{fig:level_radar}
\end{figure*}

Our motivation for designing three distinct question formats (Yes/No, Multiple-Choice, Caption Generation) is to create a more comprehensive evaluation suite that assesses a model's capabilities from multiple angles. Different formats inherently pose different challenges, testing both discriminative understanding and generative abilities. 
While the main benchmark uses different questions for each format (as shown in Figure~\ref{fig:level_radar}), here we conduct a controlled experiment to specifically isolate the impact of the question format itself.
For this analysis, we construct a small, dedicated subset of 1,000 instances where the same underlying question about a video is presented in all three formats. 
This allows for a direct comparison of model performance on the same semantic task, with only the response format varied. As shown in Table~\ref{tab:format-impact}, the results reveal a clear and consistent difficulty gradient across all tested models. 
Yes/No questions consistently yield the highest accuracy, followed by multiple-choice QA, with caption generation proving to be the most challenging format. 
This hierarchy is intuitive: a binary choice is the simplest form of discrimination, selecting from four options introduces more complex distractor-based reasoning, and generating a caption requires not only correct understanding but also precise compositional language skills. 
This finding validates our multi-format approach, confirming that it provides a progressively challenging testbed for LVMs.

\begin{table}[ht]
\centering
\resizebox{\linewidth}{!}{
\begin{tabular}{lccc}
\toprule
Model & Yes/No Acc. & Multiple-Choice Acc. & Caption Generation Acc. \\
\midrule
InterVL2 & 79.45 & 69.43 & 62.51 \\
Qwen2-VL & 83.08 & 72.93 & 65.88 \\
GPT-4o & 86.50 & 77.24 & 72.07 \\
Gemini-1.5-Pro & 88.19 & 79.60 & 75.68 \\
\bottomrule
\end{tabular}
}
\vspace{-2mm}
\caption{Performance comparison across different question formats on a subset.}
\label{tab:format-impact}
\end{table}

\subsection{Validation of the Hallucination Type Classifier}
The entire adaptive reasoning pipeline of \texttt{Dr.V-Agent} hinges on the accuracy of its initial classification step (Step 1), where GPT-4o analyzes a QA pair to determine the required level of analysis (Perceptive, Temporal, or Cognitive). 
An error in this foundational step would lead to the selection of an incorrect reasoning path, invalidating the subsequent diagnosis. Therefore, it is crucial to verify the reliability of this classifier.
To do so, we conduct a manual verification study. We randomly sample 1,000 instances from our \texttt{Dr.V-Bench} and have human experts manually label each instance with the correct hallucination level according to our taxonomy. 
We then compare these ground-truth labels against the classifications made by GPT-4o using the prompt designed for Step 1. The evaluation reveals that GPT-4o achieves an accuracy of 99.6\% (996 out of 1000 correct classifications). 
The few misclassifications occur in ambiguous cases at the boundary between temporal and cognitive reasoning. 
This near-perfect accuracy confirms that our framework can reliably and automatically direct each query to the appropriate, specialized reasoning path, ensuring the validity and efficiency of the \texttt{Dr.V-Agent} system.

\subsection{Verification of GPT’s Evaluation Reliability}
To verify the reliability of GPT-4o's evaluation in Caption Generation QA, we randomly sample 200 instances from the corresponding dataset and compare its judgments with human annotations. 
Each instance is independently reviewed by two human annotators, with a third annotator resolving conflicts when necessary. We measure agreement using accuracy and Cohen’s Kappa, which accounts for chance agreement.
GPT-4o achieves a near-perfect agreement of 98.5\%, with a Cohen's Kappa of 0.96, confirming its reliability as an evaluator for Caption Generation QA.

\subsection{Evaluating the Effectiveness of Dr.V-Agent}

The Self-PEP framework relies on an internal two-step process: (1) self-improvement, where the model generates a description to use as auxiliary context, and (2) self-explanation, where it uses an explain-then-predict approach to verify its own reasoning.

First, the model generates a video description using the prompt:  
\begin{tcolorbox}[breakable]
{``Describe the video: "}
\end{tcolorbox}

Next, the model answers the question based on both the video and the generated description:  
\begin{tcolorbox}[breakable]
{``Description: \{description\} Please provide a clear response to the question below by watching the video. If necessary, you can also use the accompanying Description to help refine your answer. Your response should be simple ``yes" or ``no". Question: \{question\} Answer the question using ``yes" or ``no": "}
\end{tcolorbox}

Finally, the model explains its reasoning and verifies the accuracy of its response before finalizing the answer:  
\begin{tcolorbox}[breakable]
{``Description: \{description\} Please offer a detailed explanation for your answer to the following question. After explaining, verify the accuracy of the information you’ve used in your explanation. Once you’ve confirmed the facts, please respond to the question with a simple ``yes" or ``no". Question: \{question\} Answer: \{predict\} Answer the question using ``yes" or ``no": "}
\end{tcolorbox}

\begin{table*}[!t]
\centering
\renewcommand{\arraystretch}{1.2}
\resizebox{\linewidth}{!}{
\begin{tabular}{lccccccccccccccc}
\toprule
\textbf{Model} & \textbf{Obj.} & \textbf{Col.} & \textbf{Num.} & \textbf{Loc.} & \textbf{SRel.} & \textbf{OCR} & \textbf{Act.} & \textbf{Atr.} & \textbf{DRel.} & \textbf{Seq.} & \textbf{Fct.} & \textbf{CnFct.} & \textbf{Cxt.} & \textbf{Knk.} & \textbf{Avg.}\\ 
\midrule
VideoChat2  & 41.88 & 38.15 & 30.14 & 64.22 & 36.20 & 30.50 & 34.06 & 28.61 & 37.77 & 28.52 & 36.68 & 33.96 & 36.35 & 30.85 & \textbf{36.28} \\
\quad {+ Self-PEP} & 49.52 & 43.60 & 35.17 & 66.81 & 45.12 & 39.10 & 37.19 & 31.04 & 45.52 & 32.70 & 43.65 & 37.22 & 43.91 & 34.68 & 42.33 (\textcolor{red}{+6.05}) \\
\quad {+ Dr.V-Agent} & 65.40 & 54.91 & 45.62 & 72.18 & 63.61 & 56.95 & 43.69 & 36.09 & 61.55 & 41.38 & 58.13 & 43.99 & 59.61 & 42.62 & \textbf{54.88} (\textcolor{red}{+18.60}) \\
\midrule
Video-ChatGPT  & 39.05 & 35.15 & 36.30 & 60.62 & 33.00 & 41.68 & 35.09 & 34.94 & 38.19 & 30.78 & 33.61 & 36.36 & 38.28 & 39.05 & \textbf{38.01} \\
\quad {+ Self-PEP} & 47.68 & 41.30 & 41.98 & 63.54 & 43.05 & 51.39 & 38.63 & 37.68 & 46.93 & 35.49 & 41.48 & 40.04 & 46.81 & 43.37 & 44.82 (\textcolor{red}{+6.81}) \\
\quad {+ Dr.V-Agent} & 58.55 & 49.05 & 49.12 & 67.22 & 55.70 & 63.62 & 43.07 & 41.14 & 57.90 & 41.43 & 51.39 & 44.68 & 57.53 & 48.81 & \textbf{53.43} (\textcolor{red}{+15.42}) \\
\midrule
Video-LLaVA  & 53.57 & 50.58 & 48.98 & 80.93 & 56.65 & 42.86 & 38.38 & 39.93 & 45.09 & 45.02 & 48.49 & 39.31 & 61.16 & 50.03 & \textbf{50.07} \\
\quad {+ Self-PEP} & 63.99 & 58.00 & 55.83 & 84.45 & 68.79 & 54.60 & 42.65 & 43.24 & 55.63 & 50.71 & 58.00 & 43.75 & 71.46 & 55.24 & 58.31 (\textcolor{red}{+8.24}) \\
\quad {+ Dr.V-Agent} & 73.68 & 64.91 & 62.19 & 87.73 & 80.07 & 65.49 & 46.61 & 46.32 & 65.44 & 56.00 & 66.83 & 47.88 & 81.02 & 60.09 & \textbf{65.96} (\textcolor{red}{+15.89}) \\
\midrule
LLaMA-VID  & 52.42 & 44.64 & 46.52 & 78.72 & 55.94 & 51.40 & 39.08 & 40.84 & 43.05 & 37.99 & 45.99 & 51.66 & 61.03 & 53.86 & \textbf{50.22} \\
\quad {+ Self-PEP} & 62.10 & 51.54 & 52.90 & 82.00 & 67.23 & 62.31 & 43.05 & 43.92 & 52.85 & 43.28 & 54.82 & 55.79 & 70.62 & 58.71 & 57.88 (\textcolor{red}{+7.66}) \\
\quad {+ Dr.V-Agent} & 72.22 & 58.77 & 59.56 & 85.42 & 79.03 & 73.70 & 47.19 & 47.14 & 63.10 & 48.81 & 64.07 & 60.12 & 80.63 & 63.78 & \textbf{65.89} (\textcolor{red}{+15.67}) \\
\midrule
LLaVA-NeXT  & 54.76 & 58.32 & 58.02 & 83.65 & 63.39 & 49.25 & 48.19 & 44.79 & 49.47 & 39.00 & 45.76 & 60.77 & 70.64 & 57.46 &\textbf{56.80} \\
\quad {+ Self-PEP} & 49.24 & 54.39 & 54.40 & 81.79 & 56.97 & 43.04 & 45.93 & 43.04 & 43.90 & 35.99 & 40.74 & 58.42 & 65.20 & 54.70 & 52.45 (\textcolor{red}{-4.35}) \\
\quad {+ Dr.V-Agent} & 76.76 & 74.01 & 72.49 & 91.09 & 89.04 & 74.01 & 57.20 & 51.79 & 71.72 & 51.02 & 65.82 & 70.15 & 92.39 & 68.48 & \textbf{74.21} (\textcolor{red}{+17.41}) \\
\midrule
PLLaVA  & 68.00 & 65.80 & 73.03 & 83.56 & 76.65 & 57.54 & 43.60 & 49.00 & 55.93 & 44.50 & 59.71 & 65.66 & 71.53 & 61.73 & \textbf{62.58} \\
\quad {+ Self-PEP} & 61.30 & 61.02 & 68.62 & 81.29 & 68.84 & 49.99 & 40.86 & 46.87 & 49.15 & 40.84 & 53.60 & 62.80 & 64.91 & 58.38 & 57.28 (\textcolor{red}{-5.30}) \\
\quad {+ Dr.V-Agent} & 81.80 & 75.63 & 82.09 & 88.23 & 92.71 & 73.07 & 49.24 & 53.38 & 69.87 & 52.03 & 72.28 & 71.54 & 85.15 & 68.63 & \textbf{73.47} (\textcolor{red}{+10.89}) \\
\midrule
InternVL2  & 73.88 & 72.95 & 72.82 & 90.93 & 76.14 & 64.20 & 60.66 & 50.96 & 64.84 & 52.83 & 63.47 & 61.22 & 74.55 & 64.38 & \textbf{67.42} \\
\quad {+ Self-PEP} & 81.52 & 78.40 & 77.85 & 93.52 & 85.06 & 72.80 & 63.79 & 53.39 & 72.59 & 57.01 & 70.44 & 64.48 & 82.11 & 68.21 & 73.47 (\textcolor{red}{+6.05}) \\
\quad {+ Dr.V-Agent} & 89.15 & 83.85 & 82.88 & 96.10 & 93.95 & 81.40 & 66.91 & 55.82 & 80.31 & 61.17 & 77.40 & 67.75 & 89.67 & 72.02 & \textbf{79.49} (\textcolor{red}{+12.07}) \\
\midrule
Qwen2-VL  & 76.57 & 76.66 & 72.84 & 90.33 & 79.31 & 73.55 & 65.72 & 55.81 & 70.55 & 64.02 & 70.01 & 68.83 & 81.21 & 71.91 & \textbf{72.67} \\
\quad {+ Self-PEP} & 80.37 & 79.37 & 75.34 & 91.61 & 83.73 & 77.83 & 67.27 & 57.02 & 74.39 & 66.10 & 73.47 & 70.45 & 84.96 & 73.81 & 75.67 (\textcolor{red}{+3.00}) \\
\quad {+ Dr.V-Agent} & 89.19 & 85.66 & 81.14 & 94.60 & 94.03 & 87.76 & 70.89 & 59.83 & 83.33 & 70.92 & 81.53 & 74.21 & 93.69 & 78.23 & \textbf{82.64} (\textcolor{red}{+9.97}) \\
\midrule
GPT-4o  & 81.25 & 81.16 & 78.03 & 93.02 & 82.81 & 78.51 & 72.58 & 60.83 & 75.20 & 68.76 & 74.33 & 73.74 & 86.03 & 75.85 & \textbf{77.29} \\
\quad {+ Self-PEP} & 87.63 & 85.70 & 82.22 & 95.18 & 90.23 & 85.69 & 75.19 & 62.86 & 81.65 & 72.24 & 80.14 & 76.46 & 92.33 & 79.04 & 82.33 (\textcolor{red}{+5.04}) \\
\quad {+ Dr.V-Agent} & 95.23 & 91.13 & 87.23 & 97.75 & 98.12 & 94.26 & 78.31 & 66.28 & 89.36 & 76.40 & 87.30 & 79.71 & 99.06 & 82.90 & \textbf{88.36} (\textcolor{red}{+11.07}) \\
\midrule
Gemini-1.5-Pro  & 83.29 & 83.55 & 80.90 & 94.04 & 84.97 & 80.34 & 75.76 & 63.51 & 77.76 & 70.54 & 77.20 & 76.46 & 88.24 & 79.01 & \textbf{79.68} \\
\quad {+ Self-PEP} & 89.67 & 88.09 & 85.09 & 96.02 & 92.39 & 87.52 & 78.37 & 65.54 & 84.21 & 74.02 & 83.01 & 79.18 & 94.34 & 82.20 & 84.72 (\textcolor{red}{+4.95}) \\
\quad {+ Dr.V-Agent} & 97.77 & 93.87 & 90.43 & 98.94 & 99.04 & 96.66 & 81.69 & 68.12 & 92.42 & 78.45 & 90.41 & 82.64 & 98.87 & 86.26 & \textbf{91.12} (\textcolor{red}{+10.84}) \\
\bottomrule
\end{tabular}
}
\vspace{-2mm}
\caption{Performance of LVMs across various hallucination types when equipped with Self-PEP or \textbf{\texttt{Dr.V-Agent}}.}
\label{improvement_full}
\end{table*}

\subsection{Qualitative Analysis and Case Studies} 
\begin{figure*}[!t]
 \centering
 \includegraphics[width=0.95\linewidth]{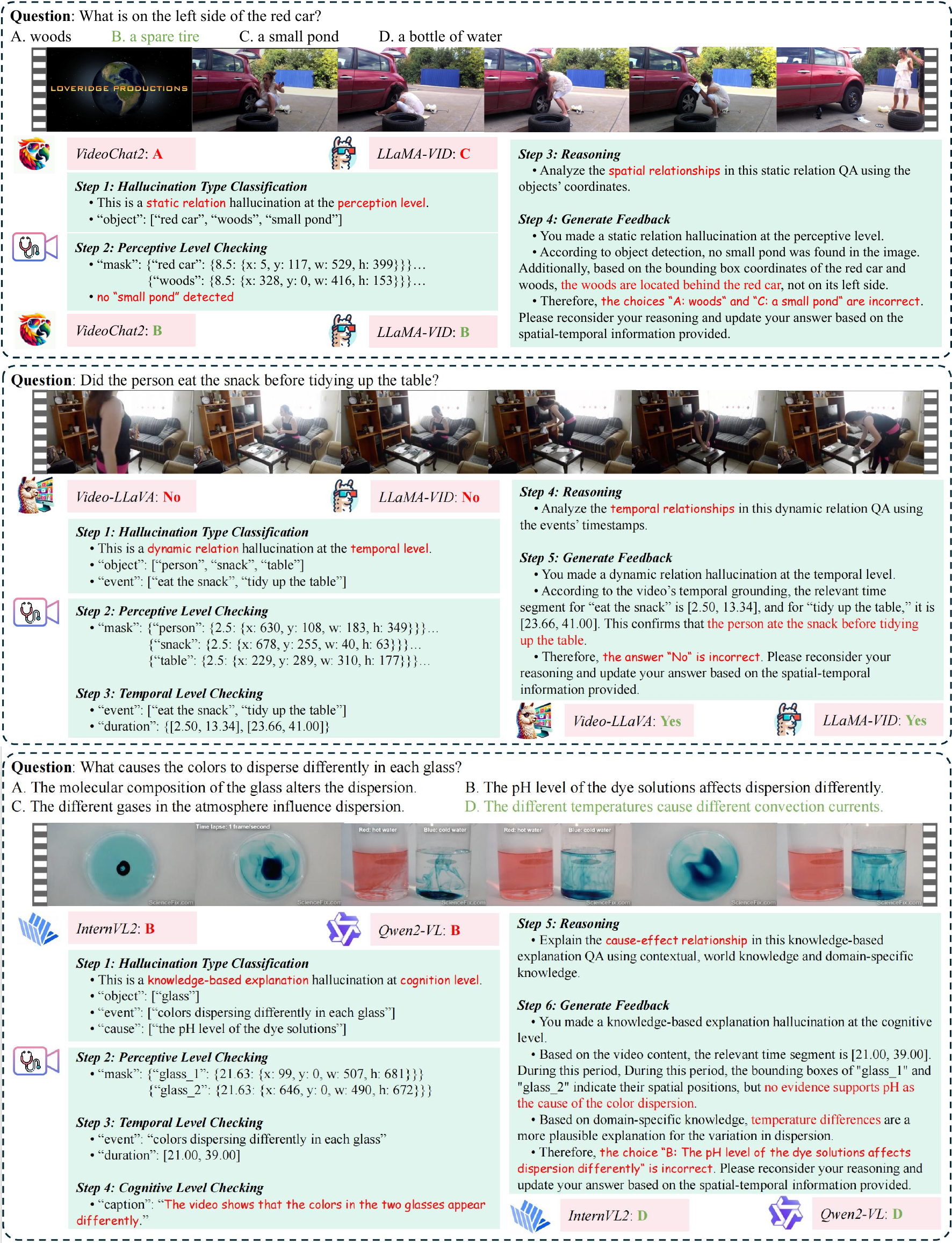}
  \vspace{-2mm}
 \caption{Qualitative examples of \textbf{\texttt{Dr.V-Agent}}'s Perceptive, Temporal, and Cognitive reasoning.}
 \label{fig:case_per_to_co}
\end{figure*}

To comprehensively evaluate Dr.V-Agent's reasoning robustness, we conduct qualitative analyses across three levels: perceptive, temporal, and cognitive. Our framework effectively mitigates hallucinations in LVMs by implementing structured evidence-based verification. As illustrated in Figure \ref{fig:case_per_to_co}, our approach ensures a multi-faceted and robust validation of the LVM's responses.

At the perceptive level, we analyze queries that require precise visual grounding of objects and their attributes. While standard LVMs frequently hallucinate the existence of target objects, Dr.V-Agent systematically cross-validates object presence and properties across multiple frames using spatial-temporal representations.
At the temporal level, we analyze event understanding and the tracking of dynamic relationships over time. Due to the limitations of common LVMs, these models often misorder events. In contrast, Dr.V-Agent's temporal grounding capability reconstructs event chronology through timestamp-aware state transitions, ensuring coherent event ordering.
At the cognitive level, we evaluate the integration of contextual and commonsense knowledge. While conventional models often generate plausible but ungrounded hypotheses, Dr.V-Agent explicitly verifies that cognitive-level conclusions remain consistent with both observed visual evidence and external knowledge bases.

\begin{figure*}[!t]
 \centering
 \includegraphics[width=1.0\linewidth]{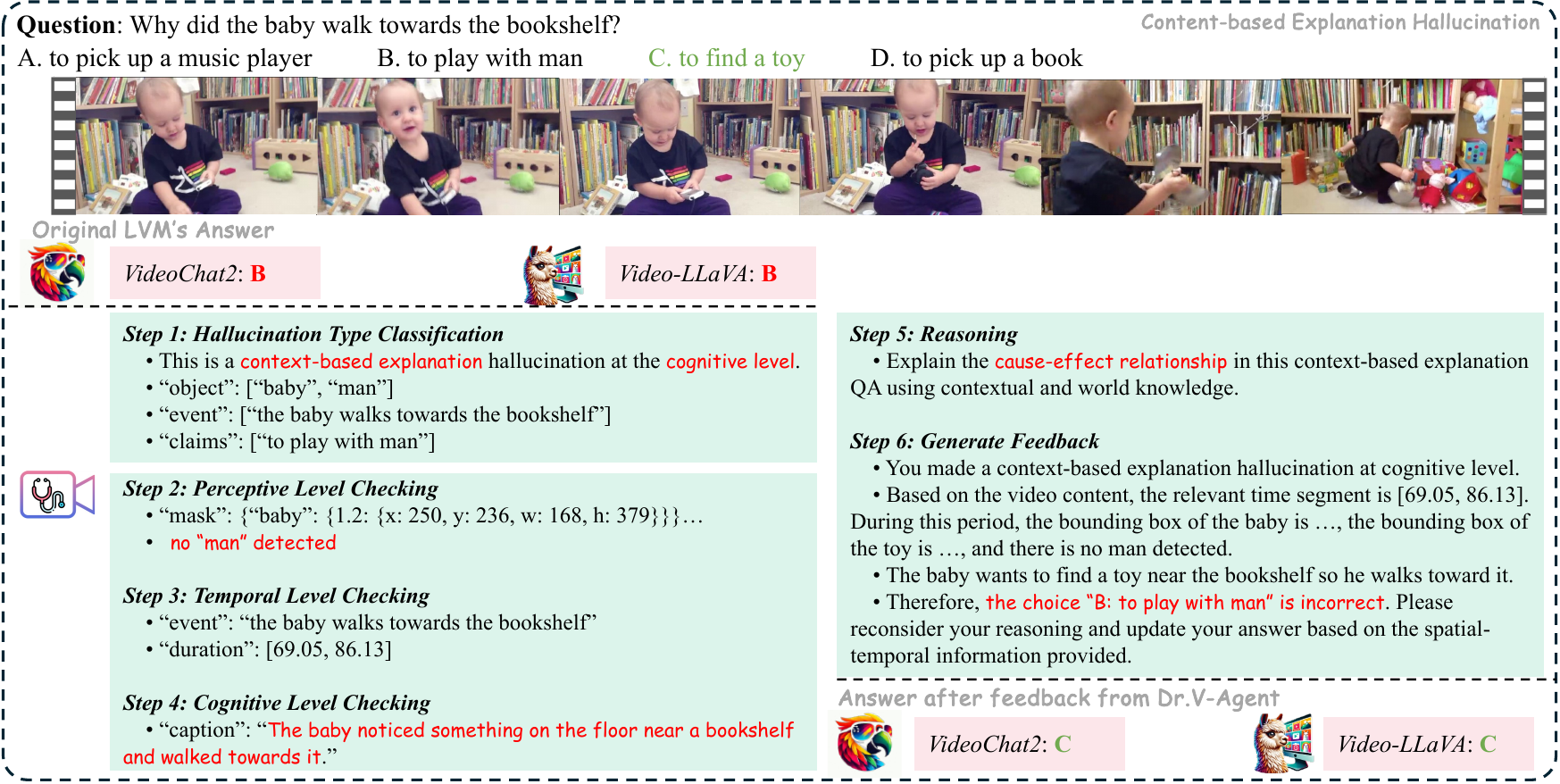}
 \vspace{-2mm}
 \caption{Visualization of a real example showcasing how Dr.V-Agent achieves successful video hallucination alleviation. The agent's feedback loop corrects the LVM's initial incorrect assessment.}
 \label{fig:casestudy_specific}
\end{figure*}

To further demonstrate the usefulness of this framework, we present a challenging case study that showcases how \texttt{Dr.V-Agent}’s feedback refines the response of LVMs. 
As illustrated in Figure \ref{fig:casestudy_specific}, the task requires the system to locate and track a target object’s spatiotemporal changes across multiple scenes and then leverage contextual knowledge to accurately infer the correct cause-and-effect relationship.
Unlike LVMs that risk hallucinating non-existent objects or motives, Dr.V-Agent confirms each claim against the actual video. It produces constructive, evidence-based feedback enabling the LVM to correct its initial response, thereby alleviating hallucination. This process is a direct application of the perceptive, temporal, and cognitive verification layers.

\end{document}